# A theory of sequence indexing and working memory in recurrent neural networks

**E. Paxon Frady[1], Denis Kleyko[2], Friedrich T. Sommer[1]**
[1]Redwood Center for Theoretical Neuroscience, U.C. Berkeley
[2]Department of Computer Science, Electrical and Space Engineering, Lulea University of Technology

**Abstract**

To accommodate structured approaches of neural computation, we propose a class of recurrent neural networks for indexing and storing sequences of symbols or analog data vectors. These networks with randomized input weights and orthogonal recurrent weights implement coding principles previously described in *vector symbolic architectures* (VSA), and leverage properties of *reservoir computing*. In general, the storage in reservoir computing is lossy and crosstalk noise limits the retrieval accuracy and information capacity. A novel theory to optimize memory performance in such networks is presented and compared with simulation experiments. The theory describes linear readout of analog data, and readout with winner-take-all error correction of symbolic data as proposed in VSA models. We find that diverse VSA models from the literature have universal performance properties, which are superior to what previous analyses predicted. Further, we propose novel VSA models with the statistically optimal Wiener filter in the readout that exhibit much higher information capacity, in particular for storing analog data.

The presented theory also applies to memory buffers, networks with gradual forgetting, which can operate on infinite data streams without memory overflow. Interestingly, we find that different forgetting mechanisms, such as attenuating recurrent weights or neural nonlinearities, produce very similar behavior if the forgetting time constants are aligned. Such models exhibit *extensive capacity* when their forgetting time constant is optimized for given noise conditions and network size. These results enable the design of new types of VSA models for the online processing of data streams.

# 1 Introduction

An important aspect of information processing is data representation. In order to access and process data, addresses or keys are required to provide a necessary context. To enable flexible contextual structure as required in cognitive reasoning, connectionist models have been proposed that represent data and keys in a high-dimensional vector space. Such models include holographic reduced representations (HRR) (Plate, 1991, 2003), and hyperdimensional computing (HDC) (Gayler, 1998; Kanerva, 2009), and will be referred to here by the umbrella term *vector symbolic architectures* (VSA; see Gayler (2003); Methods 4.1.1). VSA models have been shown to be able to solve challenging tasks of cognitive reasoning (Rinkus, 2012; Kleyko and Osipov, 2014; Gayler, 2003). VSA principles have been recently incorporated into standard neural networks for challenging machine-learning tasks (Eliasmith et al., 2012), inductive reasoning (Rasmussen and Eliasmith, 2011), and processing of temporal structure (Graves et al., 2014, 2016; Danihelka et al., 2016). Typically, VSA models offer at least two different kinds of operation, one to produce key-value bindings (also referred to as role-filler pairs), and a superposition operation that forms a working memory state containing the indexed data structures. For example, to represent a time sequence of data in a VSA, individual data points are bound to time-stamp keys and the resulting key-value pairs superposed into a working memory state.

Here, we show that input sequences can be indexed and memorized according to various existing VSA models by recurrent neural networks (RNNs) that have randomized input weights and orthonormal recurrent weights of particular properties. Conversely, this class of networks has a straight-forward computational interpretation: in each cycle, a new random key is generated, a key-value pair is formed with the new input, and the indexed input is integrated into the network state. In the VSA literature, this operation has been referred to as *trajectory association* (Plate, 1993). The memory in these networks follows principles previously described in *reservoir computing*. The idea of reservoir computing is that a neural network with fixed recurrent connectivity can exhibit a rich reservoir of dynamic internal states. An input sequence can selectively evoke these states so that an additional decoder network can extract the input history from the current network state. These models produce and retain neural representations of inputs *on the fly*, entirely without relying on previous synaptic learning as in standard models of neural memory networks (Caianiello, 1961; Little and Shaw, 1978; Hopfield, 1982; Schwenker et al., 1996; Sommer and Dayan, 1998). Models of reservoir computing include state-dependent networks (Buonomano and Merzenich, 1995), echo-state networks (Jaeger, 2002; Lukoševičius and Jaeger, 2009), liquid-state machines (Maass et al., 2002), and related network models of memory (White et al., 2004; Ganguli et al., 2008; Sussillo and Abbott, 2009). However, it is unclear how such reservoir models create representations that enables the selective readout of past input items. Leveraging the structured approach of VSAs to compute with distributed representations, we offer a novel framework for understanding reservoir computing.

# 2 Results

## 2.1 Indexing and memorizing sequences with recurrent networks

**Network model:**
We investigate how a sequence of $M$ input vectors of dimension $D$ can be indexed by pseudo-random vectors and memorized by a recurrent network with $N$ neurons (Fig. 1). The data vectors $\mathbf{a}(m) \in I\!\!R^D$ are fed into the network through a randomized, fixed input matrix $\mathbf{\Phi} \in I\!\!R^{N \times D}$. In the context of VSA, the input matrix corresponds to the *codebook*, the matrix columns contain the set of high-dimensional random vector-symbols (*hypervectors*) used in the distributed computation scheme. In addition, the neurons might also experience some independent neuronal noise $\boldsymbol{\eta}(m) \in I\!\!R^N$ with $p(\eta_i(m)) \sim \mathcal{N}(0, \sigma_\eta^2)$. Further, feedback is provided through a matrix of recurrent weights $\lambda\mathbf{W} \in I\!\!R^{N \times N}$ where $\mathbf{W}$ is orthogonal and $0 < \lambda \leq 1$. The input sequence is *encoded* into a single network state $\mathbf{x}(M) \in I\!\!R^N$ by the recurrent neural network (RNN):

$$\mathbf{x}(m) = f(\lambda\mathbf{W}\mathbf{x}(m-1) + \mathbf{\Phi}\mathbf{a}(m) + \boldsymbol{\eta}(m)) \tag{1}$$

with $f(x)$ the component-wise neural activation function.

To estimate the input $\mathbf{a}(M-K)$ entered $K$ steps ago from the network state, the *readout* is of the form:

$$\hat{\mathbf{a}}(M-K) = g(\mathbf{V}(K)^\top \mathbf{x}(M)) \tag{2}$$

where $\mathbf{V}(K) \in I\!\!R^{N \times D}$ is a linear transform to select the input that occurred $K$ time steps in the past (Fig. 1). In some models, the readout includes a nonlinearity $g(\mathbf{h})$ to produce the final output.

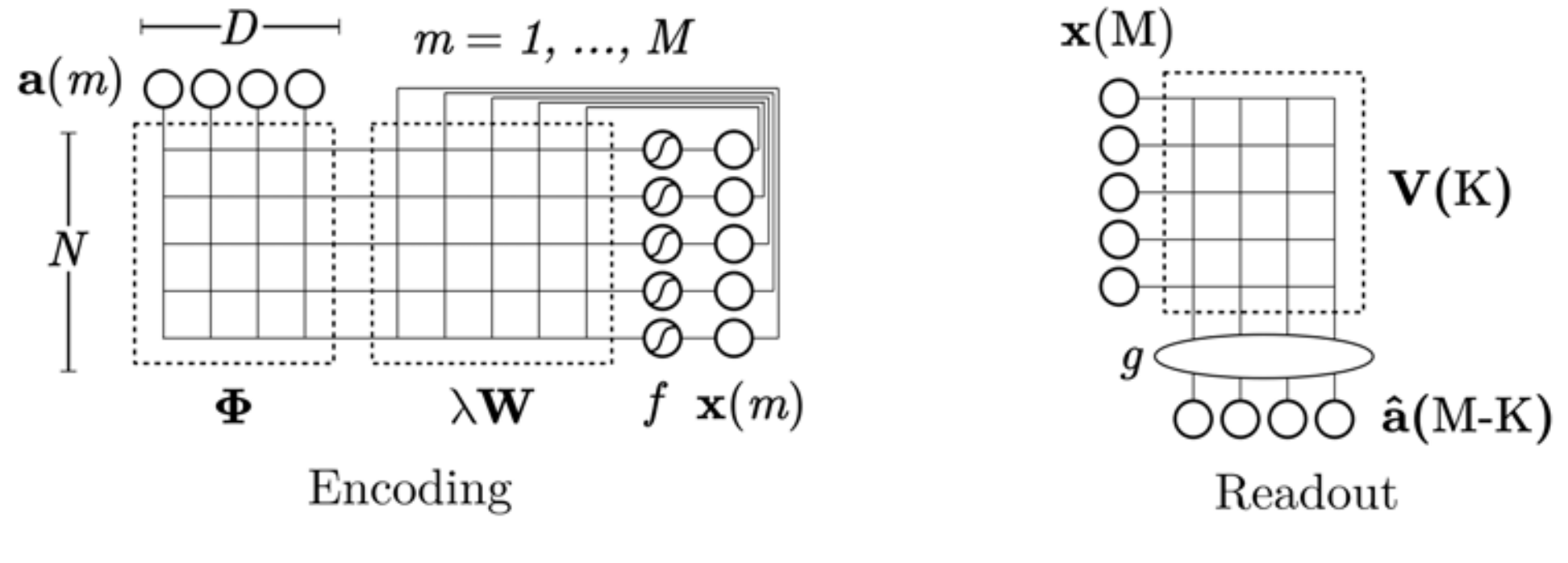

Figure 1: **Network model investigated.**

The effect of one iteration of equation (1) on the probability distribution of the network state $\mathbf{x}(m)$ is a Markov chain stochastic process, governed by the Chapman-

Kolmogorov equation (Papoulis, 1984):

$$p(\mathbf{x}(m+1)|\mathbf{a}(m)) = \int p(\mathbf{x}(m+1)|\mathbf{x}(m), \mathbf{a}(m))\ p(\mathbf{x}(m))\ d\mathbf{x}(m) \tag{3}$$

with a transition kernel $p(\mathbf{x}(m+1)|\mathbf{x}(m), \mathbf{a}(m))$, which depends on all parameters and functions in (1). Thus, to analyze the memory performance in general, one has to iterate equation (3) to obtain the distribution of the network state.

**Properties of the matrices in the encoding network:**
The analysis simplifies considerably if the input and recurrent matrix satisfy certain conditions. Specifically, we investigate networks in which the input matrix $\mathbf{\Phi}$ has i.i.d. random entries and the recurrent weight matrix $\mathbf{W}$ is orthogonal with mixing properties and long cycle length. The assumed properties of the network weights guarantee the following independence conditions of the indexing keys, which will be essential in our analysis of the network performance:

- Code vectors $\mathbf{\Phi}_d$ are composed of *identically distributed* components:

  $$p((\mathbf{\Phi}_d)_i) \sim p_{\mathbf{\Phi}}(x)\ \forall i, d \tag{4}$$

  where $p_{\mathbf{\Phi}}(x)$ is the distribution for a single component of a random code vector, and with $E_{\mathbf{\Phi}}(x)$, $V_{\mathbf{\Phi}}(x)$ being the mean and variance of $p_{\mathbf{\Phi}}(x)$, as typically defined by $E_{\mathbf{\Phi}}(\phi(x)) := \int \phi(x) p_{\mathbf{\Phi}}(x) dx$, $V_{\mathbf{\Phi}}(\phi(x)) := E_{\mathbf{\Phi}}(\phi(x)^2) - E_{\mathbf{\Phi}}(\phi(x))^2$, with $\phi(x)$ an arbitrary function.

- Components within a code vector and between code vectors are *independent*:

  $$p\left((\mathbf{\Phi}_{d'})_i, (\mathbf{\Phi}_d)_j\right) = p((\mathbf{\Phi}_{d'})_i)\ p((\mathbf{\Phi}_d)_j)\ \forall j \neq i \vee d' \neq d \tag{5}$$

- The recurrent weight matrix $\mathbf{W}$ is orthogonal and thus *preserves mean and variance* of every component of a code vector:

  $$\begin{aligned} E((\mathbf{W}\mathbf{\Phi}_d)_i) &= E((\mathbf{\Phi}_d)_i)\ \ \forall i, d \\ \mathrm{Var}((\mathbf{W}\mathbf{\Phi}_d)_i) &= \mathrm{Var}((\mathbf{\Phi}_d)_i)\ \ \forall i, d \end{aligned} \tag{6}$$

- The recurrent matrix preserves element-wise independence with a *large cycle time* (around the size of the reservoir):

  $$p((\mathbf{W}^m\mathbf{\Phi}_d)_i, (\mathbf{\Phi}_d)_i) = p((\mathbf{W}^m\mathbf{\Phi}_d)_i)\ p((\mathbf{\Phi}_d)_i)\ \ \forall i, d; m = \{1, ..., O(N)\} \tag{7}$$

The class of RNNs (1) in which the weights fulfill the properties (4)-(7) contains the neural network implementations of various VSA models. Data encoding with such networks has quite intuitive interpretation. For each input $a_d(m)$, a pseudo-random key vector is computed that indexes both the input dimension and location in the sequence, $\mathbf{W}^{M-m}\mathbf{\Phi}_d$. Each input $a_d(m)$ is multiplied with this key vector to form a new key-value pair, which is added to the memory vector $\mathbf{x}$. Each pseudo-random key defines a spatial pattern for how an input is distributed to the neurons of the network.

**Types of memories under investigation:**
*Reset memory versus memory buffer:* In the case for finite input sequence length $M$, the network is *reset* to the zero vector before the first input arrives, and the iteration is stopped after the $M$-th input has been integrated. We refer to these models as *reset memories*. In the VSA literature, the superposition operation (Plate, 1991, 2003; Gallant and Okaywe, 2013) corresponds to a reset memory, and in particular, trajectory-association (Plate, 1993). In reservoir computing, the distributed shift register (DSR) (White et al., 2004) can also be related to reset memories. In contrast, a *memory buffer* can track information from the past in a potentially infinite input stream ($M \to \infty$). Most models for reservoir computing are memory buffers (Jaeger, 2002; White et al., 2004; Ganguli et al., 2008). A memory buffer includes a mechanism for attenuating older information, which replaces the hard external reset in reset memories to avoid overload. The mechanisms of forgetting we will analyze here are contracting recurrent weights or neural nonlinearities. Our analysis links contracting weights ($\lambda$) and nonlinear activation functions ($f$) to the essential property of a memory buffer, the forgetting time constant, and we show how to optimize memory buffers to obtain extensive capacity.

*Memories for symbols versus analog input sequences:* The analysis considers data vectors $\mathbf{a}(m)$ that represent either symbolic or analog inputs. The superposition of discrete symbols in VSAs can be described by equation (1), where inputs $\mathbf{a}(m)$ are one-hot or zero vectors. A one-hot vector represents a symbol in an alphabet of size $D$. The readout of discrete symbols involves a nonlinear error correction for producing one-hot vectors as output, the winner-take-all operation $g(\mathbf{h}) = WTA(\mathbf{h})$. Typical models for reservoir computing (Jaeger, 2002; White et al., 2004) process one-dimensional analog input, and the readout is linear, $g(\mathbf{h}) = \mathbf{h}$ in equation (2). We derive the information capacity for both uniform discrete symbols and Gaussian analog inputs.

*Readout by naive regression versus full minimum mean square error regression:* Many models of reservoir computing use full optimal linear regression and set the linear transform in (2) to the Wiener filter $\mathbf{V}(K) = \mathbf{C}^{-1}\mathbf{A}(K)$, which produces the minimum mean square error (MMSE) estimate of the stored input data. Here, $\mathbf{A}(K) := \langle \mathbf{a}(M-K)\mathbf{x}(M)^\top \rangle \in I\!R^{N \times D}$ is the covariance between input and memory state, and $\mathbf{C} := \langle \mathbf{x}(M)\mathbf{x}(M)^\top \rangle \in I\!R^{N \times N}$ is the covariance matrix of the memory state. Obviously, this readout requires inverting $\mathbf{C}$. In contrast, VSA models use $\mathbf{V}(K) = c^{-1}\langle \mathbf{a}(M-K)\mathbf{x}(M)^\top \rangle = c^{-1}\mathbf{W}^K\mathbf{\Phi}$, with $c = NE_{\mathbf{\Phi}}(x^2)$ a constant, which does not require matrix inversion. Thus the readout in VSA models is computationally much simpler, but can cause reduced readout quality. We show that MMSE readout matrix can mitigate the crosstalk noise in VSA and improve readout quality in regimes where $MD \lesssim N$. This is particularly useful for the retrieval of analog input values, where the memory capacity exceeds many bits per neuron, only limited by neuronal noise.

## 2.2 Analysis of memory performance

After encoding an input sequence, the memory state $\mathbf{x}(M)$ contains information indexed with respect to the dimension $1, ..., D$ of the input vectors, and with respect to the length dimension $1, ..., M$ of the sequence. The readout of a vector component, $d$, at a particular position of the sequence, $M-K$, begins with a linear dot product operation:

$$h_d(K) := \mathbf{V}_d(K)^\top \mathbf{x}(M) \tag{8}$$

where $\mathbf{V}_d(K)$ is the $d$-th column vector of the decoding matrix $\mathbf{V}(K)$.

For readout of analog-valued input vectors, we use linear readout: $h_d(K) = \hat{a}_d(M - K) = a_d(M - K) + n_d$, where $n_d$ is decoding noise resulting from crosstalk and neuronal noise. The *signal-to-noise ratio*, $r$, of the linear readout can then be defined as:

$$r(K) := \frac{\sigma^2(a_d)}{\sigma^2(n_d)} \tag{9}$$

where we suppressed the component index $d$ and assume that the signal and noise properties are the same for all vector components.

For symbolic input, we will consider symbols from an alphabet of length $D$, which are represented by one-hot $\mathbf{a}$ vectors, that is, in each input vector there is one component $a_{d'}$ with value 1 and all other $a_d$ are 0. In this case a multivariate threshold operation can be applied after the linear readout for error correction, the winner-take-all function: $\hat{\mathbf{a}}(M - K) = WTA(\mathbf{h}(K))$.

### 2.2.1 The accuracy of retrieving discrete inputs

For symbolic inputs, we will analyze the readout of two distinct types of input sequences. In the first type, a symbol is entered in every time step and retrieval consists in *classification*, i.e., to determine which symbol was added at a particular time. The second type of input sequence can contain gaps, i.e, some positions in the sequence can be empty. If most inputs in the sequence are empty, this type of input stream has been referred to as a sparse input sequence (Ganguli and Sompolinsky, 2010). The retrieval task is then *detection* whether or not a symbol is present, and, if so, reveal its identity.

For *classification*, if $d'$ is the index of the hot component in $\mathbf{a}(M-K)$, then the readout with the winner-take-all operation is correct if in equation (8), $h_{d'}(K) > h_d(K)$ for all distractors $d \neq d'$. As we will see, under the independence conditions (4)-(7) and VSA readout, the $h_d$ readout variables are the true inputs plus Gaussian noise. The *classification accuracy*, $p_{corr}$, is:

$$
\begin{aligned}
p_{corr}(K) &= p\left(h_{d'}(K) > h_d(K)\ \forall d \neq d'\right) \\
&= \int_{-\infty}^{\infty} p(h_{d'}(K) = h)\, [p(h_d(K) < h)]^{D-1}\, dh \\
&= \int_{-\infty}^{\infty} \mathcal{N}(h'; \mu(h_{d'}), \sigma^2(h_{d'})) \left[\int_{-\infty}^{h'} \mathcal{N}(h; \mu(h_d), \sigma^2(h_d))\, dh\right]^{D-1} dh' \\
&= \int_{-\infty}^{\infty} \mathcal{N}(h'; a_{d'}, \sigma^2(n_{d'})) \left[\int_{-\infty}^{h'} \mathcal{N}(h; a_d, \sigma^2(n_d)) dh\right]^{D-1} dh'
\end{aligned}
\tag{10}
$$

For clarity in the notation of Gaussian distributions, the argument variable is added, $p(x) \sim \mathcal{N}(x; \mu, \sigma^2)$.

The Gaussian variables $h$ and $h'$ in (10) can be shifted and rescaled to yield:

$$
\begin{aligned}
p_{corr}(K) &= \int_{-\infty}^{\infty} \frac{dh}{\sqrt{2\pi}}\, e^{-\frac{1}{2}h^2} \left[\Phi\left(\frac{\sigma(h_d)}{\sigma(h_{d'})} h - \frac{\mu(h_d) - \mu(h_{d'})}{\sigma(h_{d'})}\right)\right]^{D-1} \\
&= \int_{-\infty}^{\infty} \frac{dh}{\sqrt{2\pi}}\, e^{-\frac{1}{2}h^2} \left[\Phi\left(\frac{\sigma(n_d)}{\sigma(n_{d'})} h - \frac{a_d - a_{d'}}{\sigma(n_{d'})}\right)\right]^{D-1}
\end{aligned}
\tag{11}
$$

where $\Phi$ is the Normal cumulative density function.

Further simplification can be made when $\sigma(n_{d'}) \approx \sigma(n_d)$. The *classification accuracy* then becomes:

$$
p_{corr}(s(K)) = \int_{-\infty}^{\infty} \frac{dh}{\sqrt{2\pi}}\, e^{-\frac{1}{2}h^2} \left[\Phi\left(h + s(K)\right)\right]^{D-1} \tag{12}
$$

where the *sensitivity* for detecting the hot component $d'$ from $\mathbf{h}(K)$ is defined:

$$
s(K) := \frac{\mu(h_{d'}) - \mu(h_d)}{\sigma(h_d)} = \frac{a_{d'} - a_d}{\sigma(n_d)} = \frac{1}{\sigma(n_d)} \tag{13}
$$

For *detection*, the retrieval involves two steps, to detect whether or not an input item was integrated $K$ time steps ago, and to identify which symbol if one is detected. In this case, a rejection threshold, $\theta$, is required, which governs the trade-off between the two error types: *misses* and *false positives*. If none of the components in $\mathbf{h}(K)$ exceed $\theta$, then the readout will output that no item was stored. The *detection accuracy* is given by:

$$
\begin{aligned}
p^{\theta}_{corr}(s(K)) = p\left((h_{d'}(K) > h_d(K)\ \forall d \neq d') \wedge (h_{d'}(K) \geq \theta) | \mathbf{a}(M-K) = \delta_{d=d'}\right) p_s \\
+\, p\left((h_d(K) < \theta\ \forall d) | \mathbf{a}(M-K) = 0\right) (1 - p_s)
\end{aligned}
\tag{14}
$$

where $p_s$ is the probably that $\mathbf{a}(m)$ is a nonzero signal. If the distribution of $\mathbf{h}(K)$ is close to Gaussian, the two conditional probabilities of equation (14) can be computed as follows. The accuracy, given a nonzero input was applied, can be computed analogous to equation (12):

$$
\begin{aligned}
&p\left((h_{d'}(K) > h_d(K)\ \forall d \neq d') \wedge (h_{d'}(K) \geq \theta) | \mathbf{a}(M-K) = \delta_{d=d'}\right) \\
&= \int_{(\theta-1)s(K;Mp_s)}^{\infty} \frac{dh}{\sqrt{2\pi}} e^{-\frac{1}{2}h^2} \left[\Phi\left(h + s(K;Mp_s)\right)\right]^{D-1}
\end{aligned}
\tag{15}
$$

Note that (15) is of the same form as (12), but with different integration bounds. The second conditional probability in (14), for correctly detecting a zero input, can be computed by:

$$
p\left((h_d(K) < \theta\ \forall d) | \mathbf{a}(M-K) = 0\right) = \left[\Phi\left(\theta s(K;Mp_s)\right)\right]^D \tag{16}
$$

**Special cases:**

1) As a sanity check, consider the classification accuracy (12) in the vanishing sensitivity regime, for $s \to 0$. The first factor in the integral, the Gaussian, becomes then the inner derivative of the second factor, the cumulative Gaussian raised to the $(D-1)$th power. With $s \to 0$, the integral can be solved analytically using the (inverse) chain rule to yield the correct chance value for the classification:

$$
p_{corr}(s \to 0) = \frac{1}{D} \Phi(h)^D |_{-\infty}^{\infty} = \frac{1}{D} \tag{17}
$$

2) The case $D = 1$ makes sense for detection but not for classification retrieval. This case falls under classical *signal detection theory* (Peterson et al., 1954), with $s$ being the sensitivity index. The detection accuracy (14) in this case becomes:

$$
\begin{aligned}
p_{corr}^{\theta}(s(K;Mp_s); D=1) = &\left(1 - \Phi\left((\theta-1)s(K;Mp_s)\right)\right) p_s \\
&+ \Phi\left(\theta s(K;Mp_s)\right)(1-p_s)
\end{aligned}
\tag{18}
$$

The threshold $\theta$ trades off *miss* and *false alarm* errors. Formulae (14)-(16) generalizes signal detection theory to higher-dimensional signals ($D$).

3) Consider classification retrieval (12) in the case $D = 2$. Since the (rescaled and translated) random variables $p(h_{d'}(K)) \sim \mathcal{N}(s,1)$ and $p(h_d(K)) \sim \mathcal{N}(0,1)$ (Fig. 2A) are uncorrelated, one can switch to a new Gaussian variable representing their difference: $y := h_d(K) - h_{d'}(K)$ with $p(y) \sim \mathcal{N}(-s,2)$ (Fig. 2B). Thus, for $D = 2$ one can compute (12) by just the Normal cumulative density function (and avoiding the integration):

$$
p_{corr}(s(K); D=2) = p(y < 0) = \Phi\left(\frac{s(K)}{\sqrt{2}}\right) \tag{19}
$$

The result (19) is the special case $d = 1$ of table entry "$10,010.8$" in Owen's table of normal integrals (Owen, 1980).

In general, for $D > 2$ and nonzero sensitvity, the $p_{corr}$ integral cannot be solved analytically, but can be numerically approximated to arbitrary precision (Methods Fig. 14).

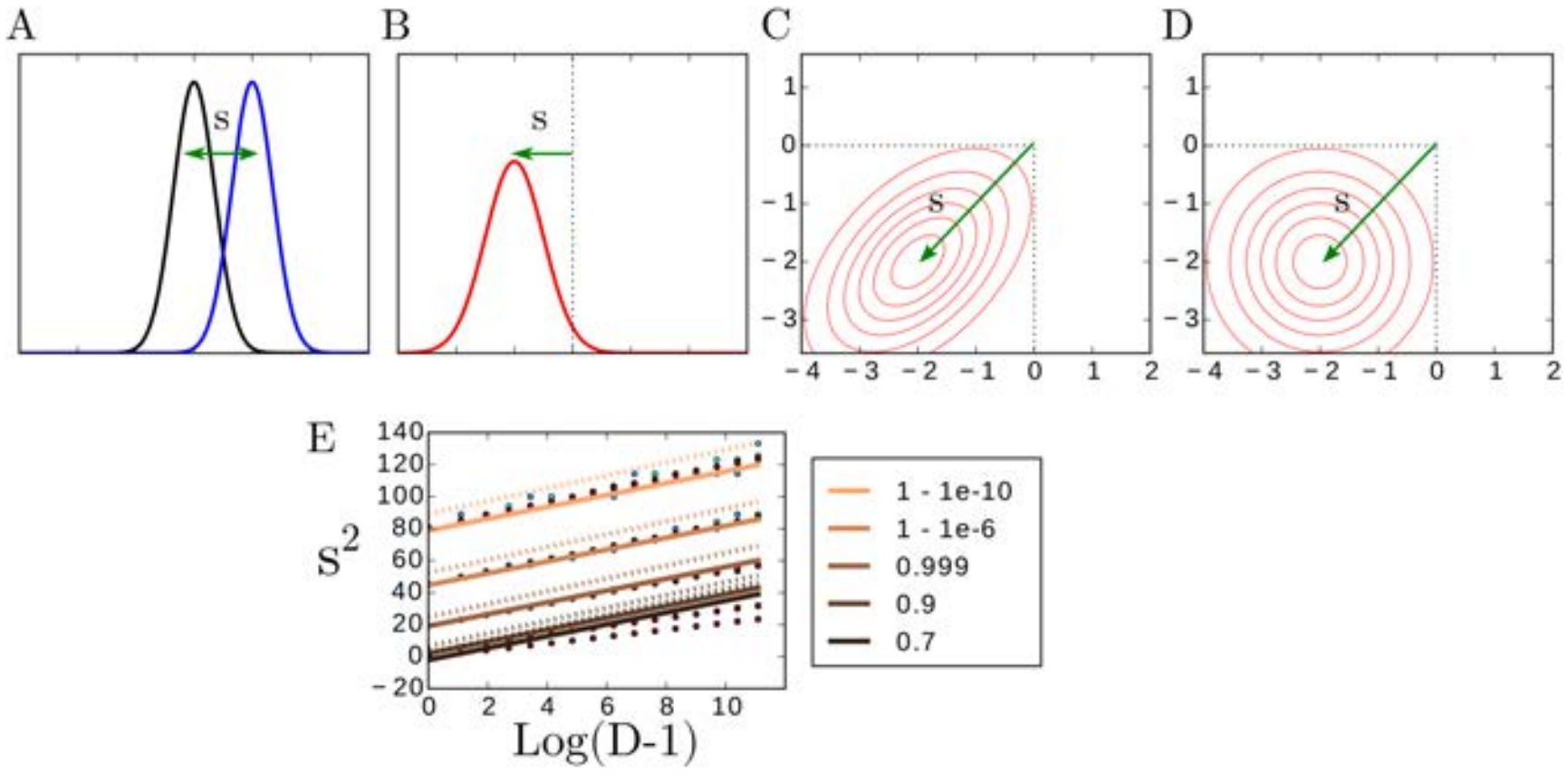

Figure 2: **Approximating the retrieval accuracy in the high-fidelity regime.** A. The retrieval is correct when the value drawn from distribution $p(h_{d'})$ (blue) exceeds the values produced by $D-1$ draws from the distribution $p(h_d)$ (black). In the shown example the sensitivity is $s = 2$. B. When $D = 2$, the two distributions can be transformed into one distribution describing the difference of both quantities, $p(h_{d'} - h_d)$. C. When $D > 2$, the $D-1$ random variables formed by such differences are correlated. Thus, in general, the multivariate cumulative Gaussian integral (20) cannot be factorized. Example shows the case $D = 3$, the integration boundaries displayed by dashed lines. D. However, for large $s$, that is, in the high-fidelity regime, the factorial approximation (21) becomes quite accurate. Panel shows again the $D = 3$ example. E. Linear relationship between the squared sensitivity and the logarithm of $D$. The numerically evaluated full theory (dots) coincides more precisely with the approximated linear theories (lines) when the accuracy is high (accuracy indicated by copper colored lines; legend). The simpler linear theory (24; dashed lines) matches the slope of the full theory but exhibits a small offset. The more elaborate linear theory (25; solid lines) provides a quite accurate fit of the full theory for high accuracy values.

### 2.2.2 Accuracy in the high-fidelity regime

Next, we derive steps to approximate the accuracy in the regime of high-fidelity recall, following the rationale of previous analyses of VSA models (Plate, 2003; Gallant and Okaywe, 2013). This work showed that the accuracy of retrieval scales linearly with the number of neurons in the network ($N$). We will compare our analysis results with those of previous analyses, and with simulation results.

Let's now try to apply to the case $D > 2$ what worked for $D = 2$ (19), that is get rid of the integral in (12) by transforming to new variables $y_d = h_d - h_{d'}$ for each of the $D-1$ distractors with $d \neq d'$. We can write $p(\mathbf{y}) \sim \mathcal{N}(-\mathbf{s}, \mathbf{\Sigma})$ with:

$$\mathbf{s} = \begin{pmatrix} s \\ s \\ \dots \end{pmatrix} \in I\!R^{D-1}, \; \mathbf{\Sigma} = \begin{pmatrix} 2 & 1 & 1 \\ 1 & 2 & \dots \\ 1 & \dots & 2 \end{pmatrix} \in I\!R^{(D-1)\times(D-1)}$$

In analogy to the $D = 2$ case (19), we can rewrite the result of (12) for $D > 2$ by:

$$p_{corr} = p(y_d < 0 \; \forall d \neq d') = \Phi_{D-1}(\mathbf{s}, \mathbf{\Sigma}) \tag{20}$$

with multivariate cumulative Gaussian, $\Phi_{D-1}(\mathbf{s}, \mathbf{\Sigma})$, see the table entry "$n0, 010.1$" in Owen's table of normal integrals (Owen, 1980).

The multivariate cumulative distribution (20) would factorize, but only for uncorrelated variables, when the covariance matrix $\mathbf{\Sigma}$ is diagonal. The difficulty with $D > 2$ is that the multiple $y_d$ variables are correlated: $\text{Cov}(y_i, y_j) = E((y_i - s)(y_j - s)) = 1$. The positive uniform off-diagonal entries in the covariance matrix means that the covariance ellipsoid of the $y_d$'s is aligned with the $(1, 1, 1, ...)$ vector and thus also with the displacement vector s (Fig. 2C). In the high signal-to-noise regime, the integration boundaries are removed from the mean and the exact shape of the distribution should not matter so much (Fig. 2D). Thus, the first step takes the *factorized approximation* (FA) to the multivariate Gaussian to approximate $p_{corr}$ in the high signal-to-noise regime:

$$p_{corr\colon FA} = \left[\Phi\left(\frac{s}{\sqrt{2}}\right)\right]^{D-1} \tag{21}$$

Note that for $s \to 0$ in the low-fidelity regime, this approximation fails; the chance probability is $1/D$ when $s \to 0$, see (17), but equation (21) yields $0.5^{D-1}$ which is much too small for $D > 2$.

For an analytic expression, an approximate formula is needed for the one-dimensional cumulative Gaussian, which is related to the complementary error function by $\Phi(x) = 1 - \frac{1}{2}\text{erfc}(x/\sqrt{2})$. A well-known exponential upper bound on the complementary error function is the *Chernoff-Rubin bound* (CR) (Chernoff, 1952). Later work Jacobs (1966); Hellman and Raviv (1970) produced a tightened version of this bound: $\text{erfc}(x) \leq B_{CR}(x) = e^{-x^2}$. Using $x = s/\sqrt{2}$, we obtain $B_{CR}(x/\sqrt{2}) = e^{-s^2/4}$, which can be inserted into (21) as the next step to yield an approximation of $p_{corr}$:

$$p_{corr\colon FA-CR} = \left[1 - \frac{1}{2}e^{-s^2/4}\right]^{D-1} \tag{22}$$

With a final approximation step, using the *local error expansion* (LEE) $e^x = 1 + x + \ldots$ when $x$ is near 0, we can set $x = -\frac{1}{2}e^{-s^2/4}$ and rewrite:

$$p_{corr\colon FA-CR-LEE} = 1 - \frac{1}{2}(D-1)e^{-s^2/4} \tag{23}$$

Solving for $s^2$ provides a simple law relating the sensitivity with the input dimension:

$$s^2 = 4\left[\ln(D-1) - \ln(2\epsilon)\right] \tag{24}$$

where $\epsilon := 1 - p_{corr}$.

The approximation (24) is quite accurate (Fig. 2E, dashed lines) but not tight. Even if (21) was tight in the high-fidelity regime, there would still be a discrepancy because the CR bound is not tight. This problem of the CR bound has been noted for long time, enticing varied efforts to derive tight bounds, usually involving more complicated multi-term expressions, e.g., (Chiani et al., 2003). Quite recently, Chang et al. (2011) studied one-term bounds of the complementary error function of the form $B(x;\alpha,\beta) := \alpha e^{-\beta x^2}$. First, they proved that there exists no parameter setting for tightening the original Chernoff-Rubin upper bound. Second, they reported a parameter range where the one-term expression becomes a lower bound: erfc$(x) \geq B(x;\alpha,\beta)$ for $x \geq 0$. The lower bound becomes the tightest with $\beta = 1.08$ and $\alpha = \sqrt{\frac{2e}{\pi}\frac{\sqrt{\beta-1}}{\beta}}$. This setting approximates the complementary error function as well as an 8-term expression derived in Chiani et al. (2003). Following Chang et al. (2011), we approximate the cumulative Gaussian with the *Chang* bound (Ch), and follow the same FA and LEE steps to derive a tighter linear fit to the true numerically evaluated integral:

$$s^2 = \frac{4}{\beta}\left[\ln(D-1) - \ln(2\epsilon) + \ln\left(\sqrt{\frac{2e}{\pi}\frac{\sqrt{\beta-1}}{\beta}}\right)\right] \tag{25}$$

with $\beta = 1.08$. This law fits the full theory in the high-fidelity regime (Fig. 2E, solid lines), but is not as accurate for smaller sensitivity values.

### 2.2.3 Information content and memory capacity

**Memory capacity for symbolic input:**

The *information content* (Feinstein, 1954) is defined as the mutual information between the true sequence and the sequence retrieved from the superposition state $\mathbf{x}(M)$. The mutual information between the individual item that was stored $K$ time steps ago ($a_{d'}$) and the item that was retrieved ($\hat{a}_d$) is given by:

$$I_{item} = D_{KL}\left(p(\hat{a}_d, a_{d'}) \mid\mid p(\hat{a}_d)p(a_{d'})\right) = \sum_{d}^{D}\sum_{d'}^{D} p(\hat{a}_d, a_{d'}) \log_2\left(\frac{p(\hat{a}_d, a_{d'})}{p(\hat{a}_d)p(a_{d'})}\right)$$

where $D_{KL}(p \mid\mid q)$ is the Kullback-Leibler divergence (Kullback and Leibler, 1951).

For discrete input sequences, because the sequence items are chosen uniformly random from the set of $D$ symbols, both the probability of a particular symbol as input and the probability of a particular symbol as the retrieved output are the same: $p(\hat{a}_d) = p(a_{d'}) = 1/D$. The $p_{corr}(s(K))$ integral evaluates the conditional probability that the output item is the same as the input item:

$$p(\hat{a}_{d'}|a_{d'}) = \frac{p(\hat{a}_{d'}, a_{d'})}{p(a_{d'})} = p_{corr}(s(K))$$

To evaluate the $p(\hat{a}_d, a_{d'})\ \forall d \neq d'$ terms, $p_{corr}(s(K))$ is needed to compute the probability of choosing the incorrect symbol given the true input. The probability that the symbol is retrieved incorrectly is $1 - p_{corr}(s(K))$, and each of the $D-1$ distractors is equally likely to be the incorrectly retrieved symbol, thus:

$$p(\hat{a}_d | a_{d'}) = \frac{p(\hat{a}_d, a_{d'})}{p(a_{d'})} = \frac{1 - p_{corr}(s(K))}{D-1} \quad \forall\, d \neq d'$$

Plugging these into the mutual information and simplifying:

$$\begin{aligned} I_{item}(p_{corr}(K)) &= p_{corr}(s(K)) \log_2 \left(p_{corr}(s(K)) D\right) \\ &\quad + (1 - p_{corr}(s(K))) \log_2 \left( \frac{D}{D-1} (1 - p_{corr}(s(K))) \right) \\ &= D_{KL} \left( \mathcal{B}_{p_{corr}(s(K))} \,||\, \mathcal{B}_{\frac{1}{D}} \right) \end{aligned} \tag{26}$$

where $\mathcal{B}_p := \{p, 1-p\}$ is the Bernoulli distribution. Note that the mutual information per item can be expressed as the Kullback-Leibler divergence between the actual recall accuracy $p_{corr}$ and the recall accuracy achieved by chance, $1/D$.

The total mutual information is the sum of the information for each item in the full sequence:

$$I_{total} = \sum_{K=1}^{M} I_{item}(p_{corr}(s(K))) = \sum_{K=1}^{M} D_{KL} \left( \mathcal{B}_{p_{corr}(s(K))} \,||\, \mathcal{B}_{\frac{1}{D}} \right) \tag{27}$$

Note that if the accuracy is the same for all items, then $I_{total} = M\ I_{item}(p_{corr})$, and by setting $p_{corr} = 1$ one obtains the entire input information: $I_{stored} = M \log_2(D)$.

**Memory capacity for analog input:**

For analog inputs, we can compute the information content if the components of input vectors are independent with Gaussian distribution, $p(a_{d'}(m)) \sim \mathcal{N}(0,1)$. In this case, distributions of the readout $p(\hat{a}_{d'})$, and the joint between input and readout $p(a_{d'}, \hat{a}_{d'})$ are also Gaussian. Therefore, the correlation between $p(a_{d'})$ and $p(\hat{a}_{d'})$ is sufficient to compute the information (Gel'fand and Yaglom, 1957), with $I_{item} = -\frac{1}{2}\log_2(1-\rho^2)$, where $\rho$ is the correlation between the input and output. There is a simple relation between the signal correlation and the SNR $r$ (9): $\rho = \sqrt{r/(r+1)}$, which gives the total information:

$$I_{total} = \frac{1}{2} \sum_{d'}^{D} \sum_{K}^{M} \log_2 \left( r(K) + 1 \right) \tag{28}$$

The *information content* of a network is $I_{total}/N$ in units bits per neuron. The *memory capacity* is then the maximum of $I_{total}/N$ (27, 28) over all parameter settings of the network. The network has *extensive memory* when $I_{total}/N$ is constant as $N$ grows large.

## 2.3 VSA indexing and readout of symbolic input sequences

In this section, we analyze the network model (1) with linear neurons, $f(\mathbf{x}) = \mathbf{x}$ and without neuronal noise. After a reset to $\mathbf{x}(0) = 0$, the network receives a sequence of $M$ discrete inputs. Each input is a one-hot vector, representing one of $D$ symbols – we will show examples with alphabet size of $D = 27$, representing the 26 English letters and the 'space' character. The readout (2) involves the matrix $\mathbf{V}(K) = c^{-1}\langle \mathbf{a}(M - K)\mathbf{x}(M)^\top \rangle = c^{-1}\mathbf{W}^K \mathbf{\Phi}$ and the winner-take-all function, with $c = E_{\mathbf{\Phi}}(x^2)N$ a scaling constant. This setting is important because, as will show in the next section, it can implement the working memory operation in various VSA models from the literature.

In this case, a sequence of inputs $\{\mathbf{a}(1), ..., \mathbf{a}(M)\}$ into the RNN (1) produces the following memory vector:

$$\mathbf{x}(M) = \sum_{m=1}^{M} \mathbf{W}^{M-m} \mathbf{\Phi} \mathbf{a}(m) \tag{29}$$

Under the conditions (4)-(7), the linear part of the readout (8) results in a sum of $N$ independent random variables:

$$h_d(K) = \sum_{i=1}^{N} \left( \mathbf{V}_d(K)^\top \mathbf{x}(M) \right)_i = c^{-1} \sum_{i=1}^{N} (\mathbf{\Phi}_d)_i (\mathbf{W}^{-K} \mathbf{x}(M))_i = c^{-1} \sum_{i=1}^{N} z_{d,i} \tag{30}$$

Note that under the conditions (4)-(7), each $z_{d,i}$ is independent and thus $h_d$ is a Gaussian by the central limit theorem for large $N$. The mean and variance of $h_d$ are given by $\mu(h_d) = c^{-1}N\mu(z_{d,i})$ and $\sigma^2(h_d) = c^{-1}N\sigma^2(z_{d,i})$.

The quantity $z_{d,i}$ in (30) can be written:

$$\begin{aligned} z_{d,i} &= (\mathbf{\Phi}_d)_i (\mathbf{W}^{-K} \mathbf{x}(M))_i \\ &= \begin{cases} (\mathbf{\Phi}_{d'})_i (\mathbf{\Phi}_{d'})_i + \sum_{m \neq (M-K)}^{M} (\mathbf{\Phi}_{d'})_i (\mathbf{W}^{M-K-m} \mathbf{\Phi}_{d'})_i & \text{if } d = d' \\ \sum_{m}^{M} (\mathbf{\Phi}_d)_i (\mathbf{W}^{M-K-m} \mathbf{\Phi}_{d'})_i & \text{otherwise} \end{cases} \end{aligned} \tag{31}$$

Given the conditions (4)-(7), the moments of $z_{d,i}$ can be computed:

$$\mu(z_{d,i}) = \begin{cases} E_{\mathbf{\Phi}}(x^2) + (M-1)E_{\mathbf{\Phi}}(x)^2 & \text{if } d = d' \\ M E_{\mathbf{\Phi}}(x)^2 & \text{otherwise} \end{cases} \tag{32}$$

$$\sigma^2(z_{d,i}) = \begin{cases} V_{\mathbf{\Phi}}(x^2) + (M-1)V_{\mathbf{\Phi}}(x)^2 & \text{if } d = d' \\ M V_{\mathbf{\Phi}}(x)^2 & \text{otherwise} \end{cases} \tag{33}$$

with $E_{\mathbf{\Phi}}(x)$, $V_{\mathbf{\Phi}}(x)$ being the mean and variance of $p_{\mathbf{\Phi}}(x)$, the distribution of a component in the codebook $\mathbf{\Phi}$, as defined by (4).

Note that with linear neurons and unitary recurrent matrix, the argument $K$ can be dropped, because there is no recency effect and all items in the sequence can be retrieved with the same accuracy.

For networks with $N$ large enough, $p(h_d(K)) \sim \mathcal{N}(c^{-1}N\mu(z_{d,i}), c^{-1}N\sigma^2(z_{d,i}))$. By inserting $\mu(h_d)$ and $\sigma(h_d)$ into (11), the accuracy then becomes:

$$p_{corr} = \int_{-\infty}^{\infty} \frac{dh}{\sqrt{2\pi}} e^{-\frac{1}{2}h^2} \times \left[ \Phi \left( \sqrt{\frac{M}{M-1+V_{\mathbf{\Phi}}(x^2)/V_{\mathbf{\Phi}}(x)^2}}\, h + \sqrt{\frac{N}{M-1+V_{\mathbf{\Phi}}(x^2)/V_{\mathbf{\Phi}}(x)^2}} \right) \right]^{D-1} \quad (34)$$

Analogous to (12), for large $M$ the expression simplifies further to:

$$p_{corr}(s) = \int_{-\infty}^{\infty} \frac{dh}{\sqrt{2\pi}} e^{-\frac{1}{2}h^2} \left[ \Phi\left(h+s\right) \right]^{D-1} \quad \text{with } s = \sqrt{\frac{N}{M}} \quad (35)$$

Interestingly, the expression (35) is independent of the statistical moments of the coding vectors and thus applies to any distribution of coding vectors $p_{\mathbf{\Phi}}(x)$ (4). Since $s$ is the ratio of $N$ to $M$, it is easy to see that this network will have *extensive capacity* when $s$ is held constant, i.e. $M = \beta N$:

$$\begin{aligned} \frac{I_{total}}{N} &= \frac{1}{N} \sum_{K=1}^{\beta N} I_{item} \left( p_{corr} \left( \sqrt{\frac{N}{\beta N}} \right) \right) \\ &= \beta I_{item} \left( p_{corr} \left( \sqrt{\frac{1}{\beta}} \right) \right) \\ &= const \end{aligned} \quad (36)$$

However, it is a complex relationship between the parameter values that actually maximizes the mutual information. We will explore this in Results 2.3.5. But first, in sections 2.3.1 to 2.3.4, we will show that $s = \sqrt{N/M}$, or a simple rescaling of it, describes the readout quality in many different VSA models.

### 2.3.1 VSA models from the literature

Many connectionist models from the literature can be directly mapped to equations (1, 2) with the settings described in the beginning of section 2.3. In the following, we will describe various VSA models and the properties of the corresponding encoding matrix $\mathbf{\Phi}$ and recurrent weight matrix $\mathbf{W}$. We will determine the moments of the code vectors required in (34) to estimate the accuracy with in the general case (for small $M$). For large $M$ values, we will show that all models perform similarly and the accuracy can predicted by the universal sensitivity formula $s = \sqrt{N/M}$.

In *hyperdimensional computing* (HDC) (Gayler, 1998; Kanerva, 2009), symbols are represented by $N$-dimensional random i.i.d. bipolar high-dimensional vectors (*hypervectors*) and referencing is performed by a permutation operation (see Methods 4.1.1). Thus, the network (1) implements encoding according to HDC when the components of the encoding matrix $\mathbf{\Phi}$ are bipolar uniform random i.i.d. variables $+1$ or $-1$, i.e., their distribution is a uniform Bernoulli distribution: $p_{\mathbf{\Phi}}(x) \sim \mathcal{B}_{0.5} : x \in \{-1, +1\}$, and $\mathbf{W}$ is a permutation matrix, a special case of a unitary matrix.

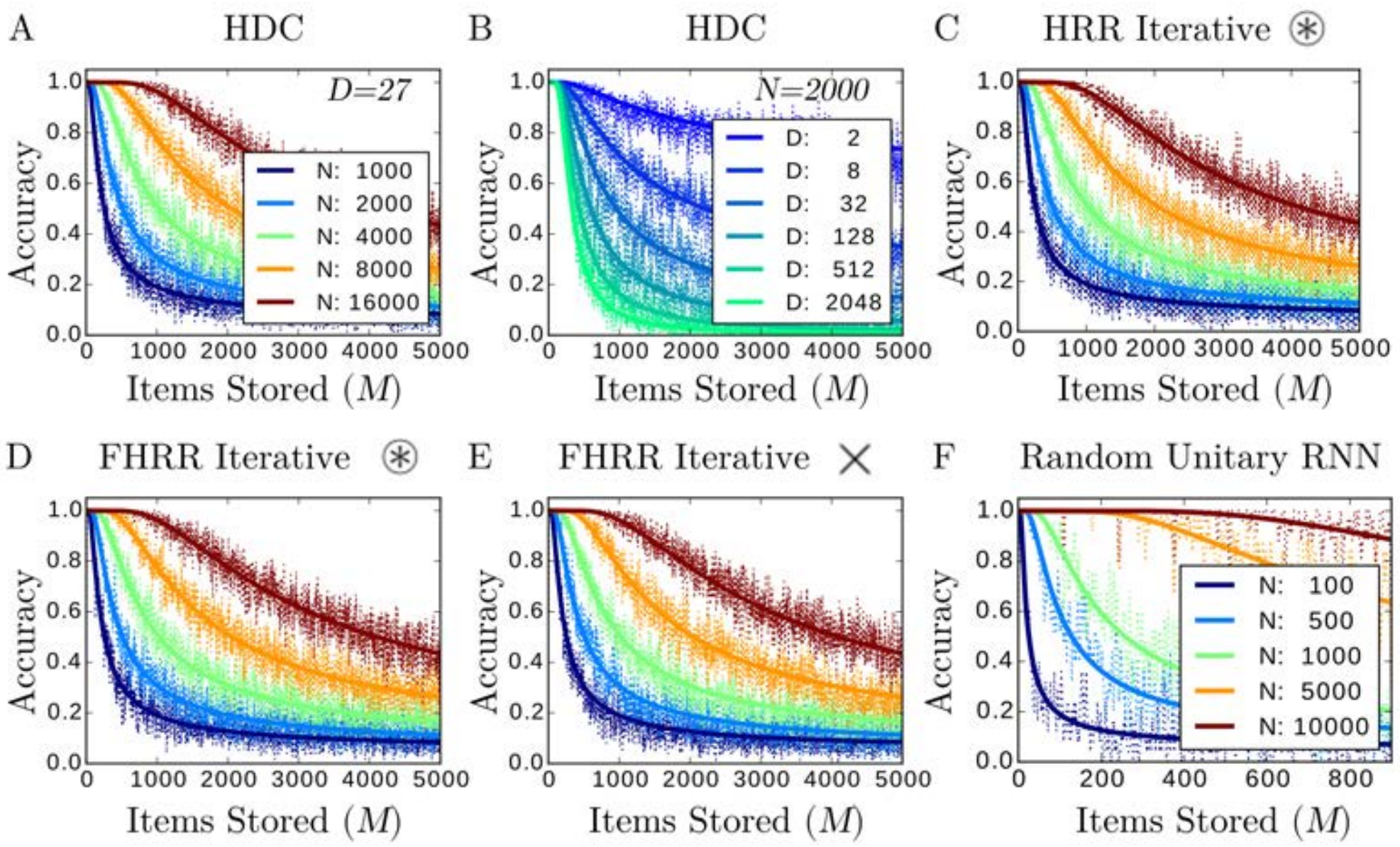

Figure 3: **Classification retrieval accuracy: theory and simulation experiments.** The theory (solid lines) matches the simulation results (dashed lines) of the sequence recall task for a variety of VSA frameworks. Alphabet length in all panels except panel B is $D = 27$. A. Accuracy of HDC code as a function of the number of stored items for different dimensions $N$ of the hypervector. B. Accuracy of HDC with different $D$ and for constant $N = 2000$. C. Accuracy of HRR code and circular convolution as binding mechanism. D. Accuracy of FHRR code and circular convolution as the binding mechanism. E. Accuracy of FHRR using multiply as the binding mechanism. F. Accuracy achieved with random encoding and random unitary recurrent matrix also performs according to the same theory.

With these settings, we can compute the moments of $z_{d,i}$. We have $E_{\mathbf{\Phi}}(x^2) = 1$, $E_{\mathbf{\Phi}}(x) = 0$, $V_{\mathbf{\Phi}}(x^2) = 0$ and $V_{\mathbf{\Phi}}(x) = 1$, which can be inserted in equation (34) to compute the retrieval accuracy. For large $M$ the retrieval accuracy can be computed using equation (35). We implemented this model and compared multiple simulation experiments to the theory. The theory fits the simulations precisely for all parameter settings of $N$, $D$ and $M$ (Fig. 3A, B).

In *holographic reduced representation* (HRR) (Plate, 1993, 2003), symbols are represented by vectors drawn from a Gaussian distribution with variance $1/N$: $p_{\mathbf{\Phi}}(x) \sim \mathcal{N}(0, 1/N)$. The binding operation is performed by circular convolution and trajectory association can be implemented by binding each input symbol to successive convolutional powers of a random *key* vector, $\mathbf{w}$. According to Plate (1995), the circular convo-lution operation can be transformed into an equivalent matrix multiply for a fixed vector by forming the circulant matrix from the vector (i.e. $\mathbf{w} \circledast \mathbf{\Phi}_d = \mathbf{W}\mathbf{\Phi}_d$). This matrix has elements $W_{ij} = w_{(i-j)\%N}$ (where the subscripts on $\mathbf{w}$ a are interpreted modulo $N$). If $||\mathbf{w}|| = 1$, the corresponding matrix is unitary. Thus, HRR trajectory association can be implemented by an RNN with a recurrent circulant matrix and encoding matrix with entries drawn from a normal distribution. The analysis described for HDC carries over to HRR and the error probabilities can be computed through the statistics of $z_{d,i}$, with $E_{\mathbf{\Phi}}(x) = 0$, $E_{\mathbf{\Phi}}(x^2) = 1/N$ giving $\mu(z_{d,i}) = (1/N)\delta_{d=d'}$, and with $V_{\mathbf{\Phi}}(x)^2 = 1/N$, $V_{\mathbf{\Phi}}(x^2) = 2/N$ giving $\sigma^2(z_{d,i}) = (M + \delta_{d=d'})/N$. We compare simulations of HRR to the theoretical results in Fig. 3C.

The *Fourier holographic reduced representation* (FHRR) (Plate, 2003) framework uses complex hypervectors as symbols, where components lay on the complex unit circle and have random phases: $(\mathbf{\Phi}_d)_i = e^{i\phi}$, with a phase angle drawn from the uniform distribution $p(\phi) \sim \mathcal{U}(0, 2\pi)$. The network implementation uses complex vectors of a dimension of $N/2$. Since each vector component is complex, there are $N$ numbers to represent (one for the real part and one for the imaginary part (Danihelka et al., 2016)). The first $N/2$ rows of the input matrix $\mathbf{\Phi}$ act on the real parts, the second $N/2$ act on the imaginary part. Trajectory-association can be performed with a random vector with $N/2$ complex elements acting as the key, raising the key to successive powers, and binding this with each input sequentially. In FHRR, both element-wise multiply or circular convolution can be used as the binding operation, and trajectory association can be performed to encode the letter sequence with either mechanism (see Methods 4.1.2 for further details). These are equivalent to an RNN with the diagonal of $\mathbf{W}$ as the key vector or as $\mathbf{W}$ being the circulant matrix of the key vector.

Given that each draw from $\mathcal{C}$ is a unitary complex number $z = (cos(\phi), sin(\phi))$ with $p(\phi) \sim \mathcal{U}(0, 2\pi)$, the statistics of $z_{d,i}$ are given by $E_{\mathbf{\Phi}}(x^2) = E(cos^2(\phi)) = 1/2$, $[E_{\mathbf{\Phi}}(x)]^2 = E(cos(\phi))^2 = 0$, giving $\mu(z_{d,i}) = \delta_{d=d'}/2$. For the variance, let $z_1 = (cos(\phi_1), sin(\phi_1))$ and $z_2 = (cos(\phi_2), sin(\phi_2))$. Then $z_1^\top z_2 = \cos(\phi_1)\cos(\phi_2) + \sin(\phi_1)\sin(\phi_2) = cos(\phi_1 - \phi_2)$. Let $\phi_* = \phi_1 - \phi_2$, it is easy to see that it also the case that $p(\phi_*) \sim \mathcal{U}(0, 2\pi)$. Therefore, $V_{\mathbf{\Phi}}(x)^2 = Var(cos(\phi_*))^2 = 1/4$ and $V_{\mathbf{\Phi}}(x^2) = 0$ giving $\sigma^2(z_{d,i}) = (M - \delta_{d=d'})/4$. Again we simulate such networks and compare to the theoretical results (Fig. 3D, E).

A random unitary matrix acting as a binding mechanism has also been proposed in the *matrix binding with additive terms* framework (MBAT) (Gallant and Okaywe, 2013). Our theory also applies to equivalent RNNs with random unitary recurrent matrices (created by QR decomposition of random Gaussian matrix), with the same $s = \sqrt{N/M}$ (Fig. 3F). Picking an encoding matrix $\mathbf{\Phi}$ and unitary recurrent matrix $\mathbf{W}$ at random satisfies the required assumptions (4)-(7) with high probability when $N$ is large.

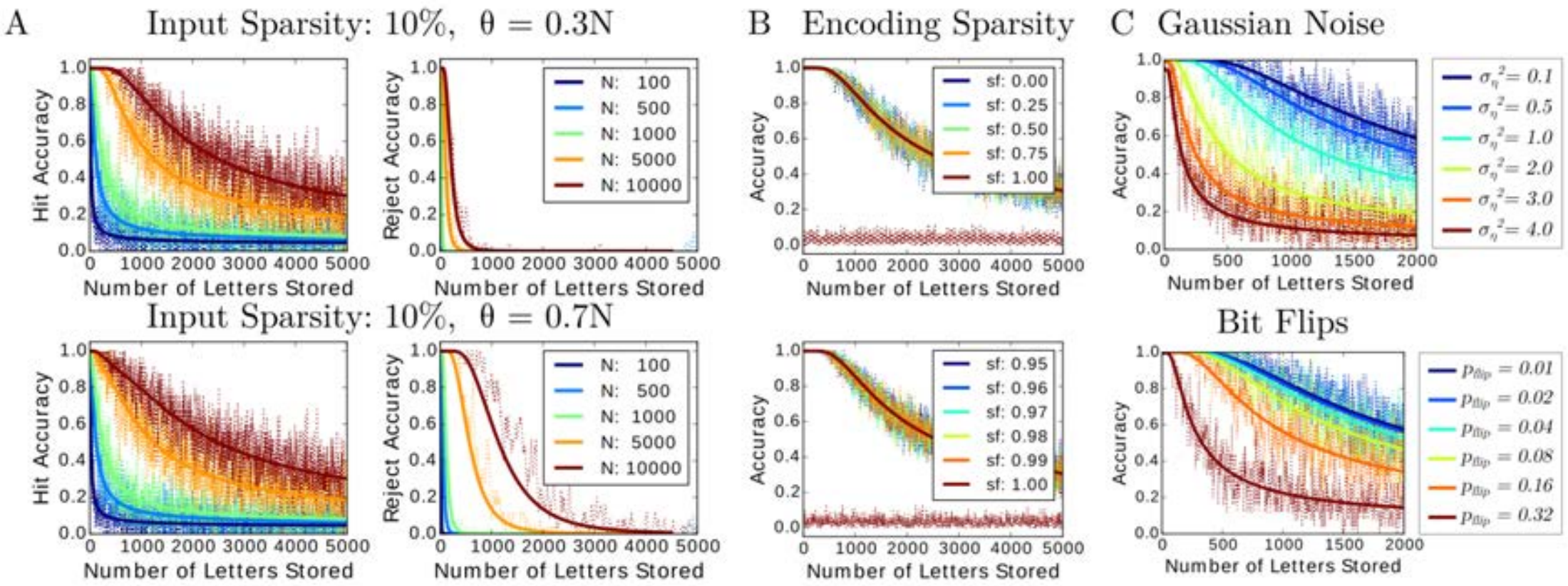

Figure 4: **Detection retrieval accuracy, encoding sparsity and noise.** A. Retrieval of a sparse input sequence ($p_s = 0.9$, $10\%$ chance for a zero vector). The hit and correct rejection performance for simulated networks (dashed lines) with different detection thresholds matches the theory (solid lines) – a rejection is produced if $h_d < \theta \; \forall d$. B. Performance is not affected by the level of encoding sparsity until catastrophic failing when all elements are $0$. C. Simulations (dashed lines) match theory (solid lines) for networks corrupted by Gaussian noise (top) and random bit-flips (bottom).

### 2.3.2 Sparse input sequences

We next analyze detection retrieval of sparse input sequences, in which the input data vector $\mathbf{a}(m)$ is nonzero only with some probability $p_s$. The readout must first decide whether an input was present, and determine its identity if present. With a random input matrix, linear neurons and a unitary recurrent matrix, the sensitivity is $s = \sqrt{N/(Mp_s)}$. The crosstalk noise only increments when the input $\mathbf{a}(m)$ generates a one-hot vector. The threshold setting trades off hit and correct rejection accuracy (miss and false positive error). We illustrate this in Fig. 4A using equations (14)-(16) describing retrieval accuracy. The readout performance for sparse data sequences depends only on the product $Mp_s$. Thus, it appears possible to recover memory items with high sensitivity even for large sequence lengths $M > N$, if $p_s$ is very small. However, our theoretical result requires assumptions (4)-(7), which break down for extremely long sequences. The result does not account for this breakdown, and is optimistic for this scenario. Previous results that consider such extremely sparse input sequences have used the methods of *compressed sensing* and *sparse inference* (Ganguli and Sompolinsky, 2010; Charles et al., 2014, 2017), and show that recovering sparse input sequences with $M > N$ is possible.

#### 2.3.3 Sparse codebook

Often neural activity is characterized as sparse and some VSA models utilize sparse codebooks. Several studies point to sparsity as an advantageous coding strategy for connectionist models (Rachkovskij, 2001). Sparse codes can be studied within our framework (1) with a sparse input matrix – i.e. a random matrix in which elements are zeroed out with a probability referred to as the *sparseness factor*, $sf$. Sparsity in the codebook affects both signal and the noise equally, and cancels out to produce the same sensitivity, $s = \sqrt{N/M}$, as with a non-sparse input matrix. Thus, sparsity essentially has no effect on the capacity of the memory, up to the catastrophic point of sparsity where entire columns in the input matrix become zero (Fig. 4B).

#### 2.3.4 Neuronal noise

Here, we consider the case where each neuron experiences i.i.d. Gaussian noise in each time step in addition to the data input. The effect of the noise depends on the ratio between noise variance and the variance of a component in the code vectors $V_{\mathbf{\Phi}}$. The sensitivity with neuronal noise is:

$$s = \sqrt{\frac{N}{M(1 + \sigma_\eta^2/V_{\mathbf{\Phi}})}} \tag{37}$$

Noise accumulation only scales $s$ by a constant factor.

There are other ways to model noise in the network. For the case where there is only white noise added during the retrieval operation, it is easy to see that this noise will be added to the variance of $z_{d,i}$, giving $s = \sqrt{N/(M + \sigma_\eta^2/V_{\mathbf{\Phi}})}$. If the noise was instead like a bit-flip in readout hypervector, with the probability of bit-flip $p_f$, then this gives $s = \sqrt{\frac{N(1-2p_f)^2}{M+2p_f}}$. Finally, with these derivations of $s$ and (12), the empirical performance of simulated neural networks is matched (Fig. 4C).

#### 2.3.5 Memory capacity of VSAs with symbolic input

The original estimate for the capacity of distributed random codes (Plate, 1993) considered a slightly different setup (see Methods 4.2.2), but follows similar ideas as the FA-CR-LEE high-fidelity approximation (24) and we reformulated the Plate (1993) derivation to compare with our analysis. This work first showed that random indexing has linear extensive capacity and that the memory capacity is at least 0.16 bits per neuron. Figure 5A compares the approximations (21, 23, 24) with the true numerically evaluated integral (12). These approximations are good in the high-fidelity regime, where $p_{corr}$ is near 1, but underestimate the performance in the low-fidelity regime.

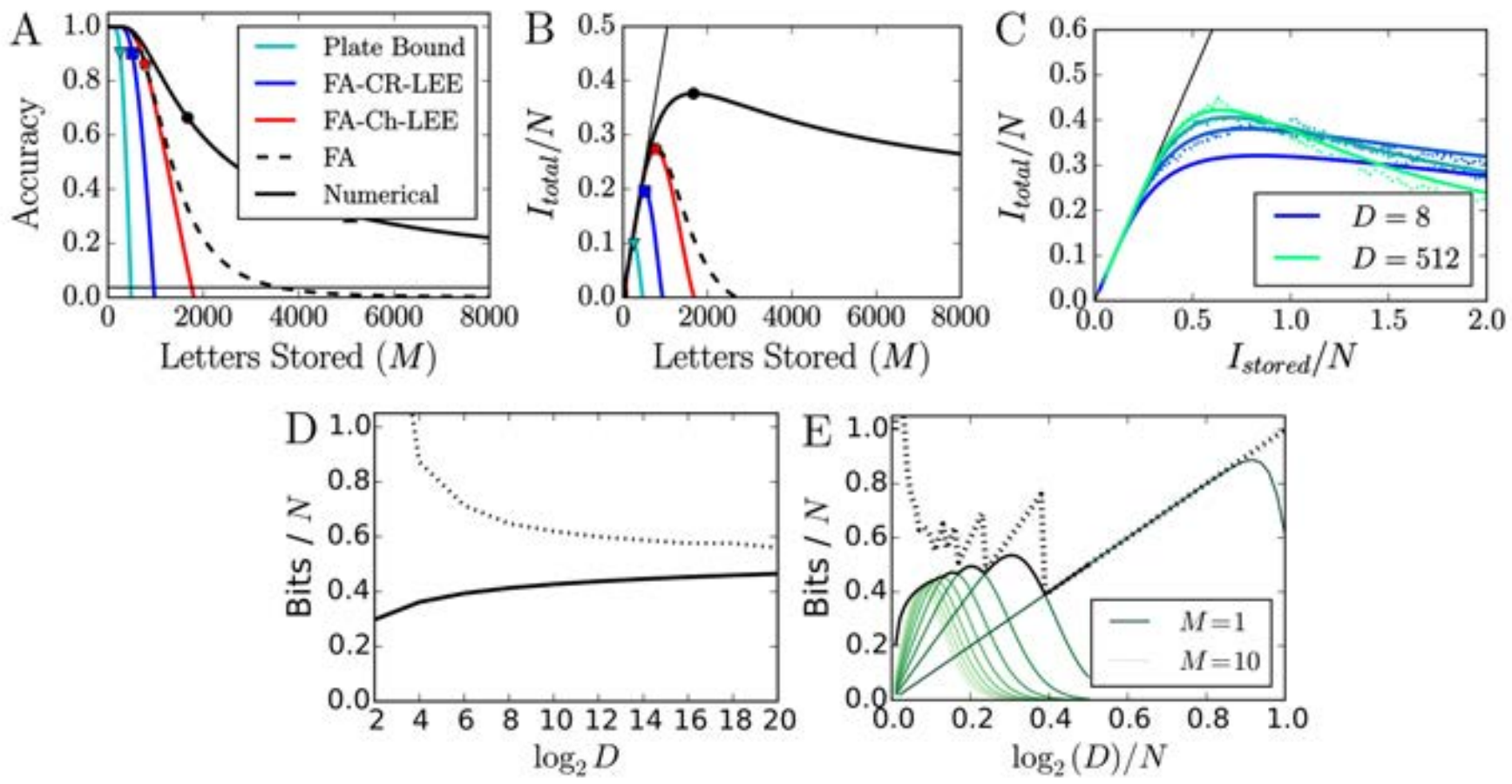

Figure 5: **Information content and memory capacity.** A. Approximations of retrieval accuracy derived in Results 2.2.2 and Plate (1993) are compared to the numerically evaluated accuracy ($p_{corr}$). The approximations underestimate the accuracy in the low fidelity regime ($D = 27$, $N = 10,000$). B. The total information content retrieved, and memory capacity (solid points). High-fidelity retrieval recovers nearly all of the stored information (thin black line, $I_{stored} = M \log_2 D$ (27)), but the true memory capacity is somewhat into the low-fidelity regime. C. Retrieved information measured in simulations (dashed lines) compared to the predictions of the theory (solid lines). The memory capacity is dependent on $D$. D. Memory capacity as a function of $D$ (solid line) and information of the input sequence at retrieval maximum ($I_{stored}$, dashed). E. Maximum information retrieved (solid black line) and total information stored ($I_{stored}$, dashed) where $D$ is a significant fraction of $2^N$ ($N = 100$). The retrieved information for fixed sequence lengths $M = \{1, ..., 10\}$ are plotted (green lines of different shades). For $M = 1$, retrieved and stored information come close; with larger $M$, the gap grows.

With the relationship $s = \sqrt{N/M}$, the information contained in the activity state $\mathbf{x}(M)$ can be calculated. We compare the total information (27) based on the approximations with the true information content determined by numeric evaluation of $p_{corr}$ (Fig. 5B). In the linear scenario with unitary weight matrix, $p_{corr}$ has no dependence on $K$, and so the total information in this case is simply $I_{total} = M I_{item}$ (27).

In the high fidelity regime with $p_{corr} = 1 - \epsilon$ and small $\epsilon$, we can estimate with (24):

$$\frac{I_{total}}{N} = \frac{M}{N} D_{KL}\left(\mathcal{B}_{1-\epsilon} \,||\, \mathcal{B}_{\frac{1}{D}}\right) \approx \frac{M \log(D)}{N \log(2)} \approx \frac{\log(D)}{4\log(2)(\log(D-1) - \log(2\epsilon))} \quad (38)$$

We can see that for any fixed, finite $D$ the information per neuron depends on the admitted error $\epsilon$ and vanishes for $\epsilon \to 0$. If the alphabet size $D$ is growing with $N$, and for fixed small error $\epsilon$, the asymptotic capacity value is $1/(4 \ln 2) = 0.36$. Our best high-fidelity approximation (25), increases the total capacity bound above previous estimates to $0.39$.

Results for the case of a finite moderate-sized alphabet size ($D = 27$) are shown in

Fig. 5B. The novel and most accurate high-fidelity approximation (25) predicts 0.27 bits per neuron, the simpler high-fidelity approximations substantially underestimate the capacity.

Importantly, however, our full theory shows that the true information maximum lies outside the high-fidelity regime. The maximum capacity for $D = 27$ is nearly 0.4 bits per neuron with (Fig. 5B, black circle). Thus, the achievable capacity is about four times larger than previous analysis suggested.

In a wide range of $D$, our full theory precisely predicts the empirically measured total information in simulations of the random sequence recall task (Fig. 5C). The total information per neuron scales linearly with the number of neurons, and the maximum amount of information per element that can be stored in the network is dependent upon $D$. The memory capacity increases with $D$, reaching over 0.5 bits per neuron for large $D$ (Fig. 5D, solid line).

**Capacity without superposition:**
As the alphabet size, $D$, grows super-linear in $N$ (approaching $2^N$), one needs to reduce $M$ in order to maximize the memory capacity (Fig. 5E, dashed line). The theory breaks down when there is no superposition, i.e. when $M = 1$. This case is different because there is no crosstalk, but for very large $D$ and randomly generated code vectors, *collisions* arise, another source of retrieval errors. Collisions are coincidental duplication of code vectors. The theory presented so far can describe effects of crosstalk but not of collisions.

For the sake of completeness, we briefly address the case of $M = 1$ and very large $D$. This case without crosstalk shows that it is possible to achieve the theoretically optimal capacity of 1 bit per neuron, and that crosstalk immediately limits the achievable capacity. If code vectors are drawn i.i.d. with a uniform Bernoulli distribution $p_{\Phi}(x) \sim \mathcal{B}_{0.5} : x \in \{-1, +1\}$, then the probability of accurately identifying the correct codeword is:

$$p_{corr}^{M=1} = \sum_{C} p_C/(C + 1) \tag{39}$$

where $p_C$ is the probability of a vector having collisions with $C$ other vectors in the codebook of $D$ vectors, which can be found based on the binomial distribution:

$$p_C = \binom{D}{C} q^C (1 - q)^{D-C} \tag{40}$$

where $q = 1/2^N$ is the likelihood of a pair of vectors colliding. The collisions reduce the accuracy $p_{corr}^{M=1}$ to $(1 - 1/e) \approx 0.63$ for $D = 2^N$ in the limit $N \to \infty$. However, this reduction does not affect the asymptotic capacity. It is $I_{total}/N \to 1$ bits per neuron as $N \to \infty$, the same as for a codebook without collisions, see Methods 4.4.3. The effects of collisions at finite sizes $N$, can be seen in Fig. 5E and Methods Fig. 18A.

In the presence of superposition, that is, for $M > 1$, the crosstalk noise becomes immediately the limiting factor for memory capacity. This is shown in Fig. 5E, for a small network of 100 neurons. For $M = 2$, the memory capacity drops to around 0.5 bits per neuron and decreases to about 0.2 bits per neuron for large $M$-values (5E, black line). The capacity curves for fixed values of $M$ (5E, green lines) show the effect of crosstalk noise, which increases as more items are superposed (as $M$ increases). For $M = 1$ (5E, dark green line), equations (39) and (27) can be evaluated as $D$ grows to $2^N$.

## 2.4 Indexed memory buffers with symbolic input

With the linear encoding network described in the previous section, there is no *recency effect*, the readout of the most recent input stored is just as accurate as readout of the earliest input stored. In contrast, a network with a contracting recurrent weight matrix (Results 2.4.1) or nonlinear neural activation functions (Results 2.4.2, and 2.4.3) will have a recency effect. In this section, the theory will be developed to describe memory in encoding networks (1) with recency effects. We juxtapose the performance of networks with recency effects in our two memory tasks, reset memory and memory buffer. We show that reset memories yield optimal capacity in the absence of any recency effect (Lim and Goldman, 2011). However, recency effects can avoid catastrophic forgetting when the input sequence is infinite. Through the recency effect, sequence items presented further back in time will be attenuated and eventually forgotten. Thus, the recency effect is a soft substitution of resetting the memory activity before input of interest is entered. Further, we show that contracting recurrent weights and saturating neural activation functions have very similar behavior if their forgetting time constants $\tau$ are aligned (Results 2.4.4). Last, we optimize the parameters of memory buffers for high memory capacity, and show that *extensive capacity* (Ganguli et al., 2008) is achievable even in the presence of neuronal noise, by keeping the forgetting time constant proportional to $N$.

### 2.4.1 Linear neurons with contracting recurrent weights

**Reset Memory:**
Consider a network of linear neurons in which the attenuation factor $0 < \lambda < 1$ contracts the network activity in each time step. After a sequence of $M$ input symbols has been applied, the variance of $z_{d,i}$ is $\left(\frac{1-\lambda^{2M}}{1-\lambda^2}\right) V_{\mathbf{\Phi}}$, and the signal decays exponentially with $\lambda^K E_{\mathbf{\Phi}}(x^2)$. The sensitivity for recalling the input that was added $K$ time steps ago is:

$$s(K) = \lambda^K \sqrt{\frac{N(1-\lambda^2)}{1-\lambda^{2M}}} \tag{41}$$

Thus, the sensitivity decays exponentially as $K$ increases, and the highest retrieval accuracy is from the most recent item stored in memory. The accuracy (Fig. 6A1) and information per item (Fig. 6B1) based on this formula for $s(K)$ shows the interde-pendence between the total sequence length ($M$) and the lookback distance ($K$) in the history.

Equation (41) is monotonically increasing as $\lambda$ increases, and thus to maximize the sensitivity for the $K$-th element in history given a finite set of $M$ stored tokens, we would want to maximize $\lambda$, or have the weight matrix remain unitary with $\lambda = 1$ [1]. The memory capacity is maximized as $\lambda \to 1$ when $M$ is finite (Fig. 6C1) and as $D$ grows large (Fig. 6D1), and there is no benefit of contracting weights in reset memories.

**Memory Buffer:**
For an infinite stream of inputs, $M \to \infty$, the setting $\lambda = 1$ results in catastrophic forgetting. However with $\lambda < 1$, the memory can operate even in this case, because past signals fade away and make room for storing new inputs. These networks with $\lambda < 1$, $f(x) = x$, and $\mathbf{W}$ unitary have been denoted *normal networks* (White et al., 2004).

The value of $\lambda$ affects both the signal and the crosstalk noise. For large $M$, the noise variance is bounded by $\frac{1}{1-\lambda^2}$, and the network reaches its *saturated* equilibrium state. The sensitivity for the $K$-th element back in time from the saturated state is:

$$s(K) = \lambda^K \sqrt{N(1-\lambda^2)} \tag{42}$$

The theory (42) and (12) predicts the performance of simulated networks with contracting recurrent weights that store a sequence of symbols with $M >> N$ via trajectory association (Fig. 6A2, solid lines) for different $\lambda$ (Fig. 6A2, dashed lines). The information per item retrieved (Fig. 6B2) and the total information (Fig. 6C2) for different values of $\lambda$ shows the trade-off between fidelity and duration of storage. There is an ideal $\lambda$ value that maximizes the memory capacity with $M \to \infty$ for given $N$ and $D$. This ideal $\lambda$ value differs depending on the alphabet size ($D$; Fig. 6D2). For larger alphabets (meaning more bits per symbol), the inputs should be forgotten more quickly and memorizing a shorter history optimizes the memory capacity (Fig. 6E). The values of $\lambda$ that maximize the memory capacity were computed numerically, they drop with increasing $N$ and $D$ (Fig. 6F).

### 2.4.2 Neurons with clipped-linear transfer function

Squashing nonlinear neural activation functions induce a recency effect, similar to contracting recurrent weights. Consider equation (1) with $\lambda = 1$ and the clipped-linear activation function, $f(x) = f_\kappa(x)$, in which the absolute value of the activity of neu-

[1] If we allowed $\lambda$ to be larger than 1, then memories from the past would grow in magnitude exponentially – this would mean higher SNR for more distant memories at the cost of lower SNR for recent memories (this would cause the network to explode, however normalization could be used.)

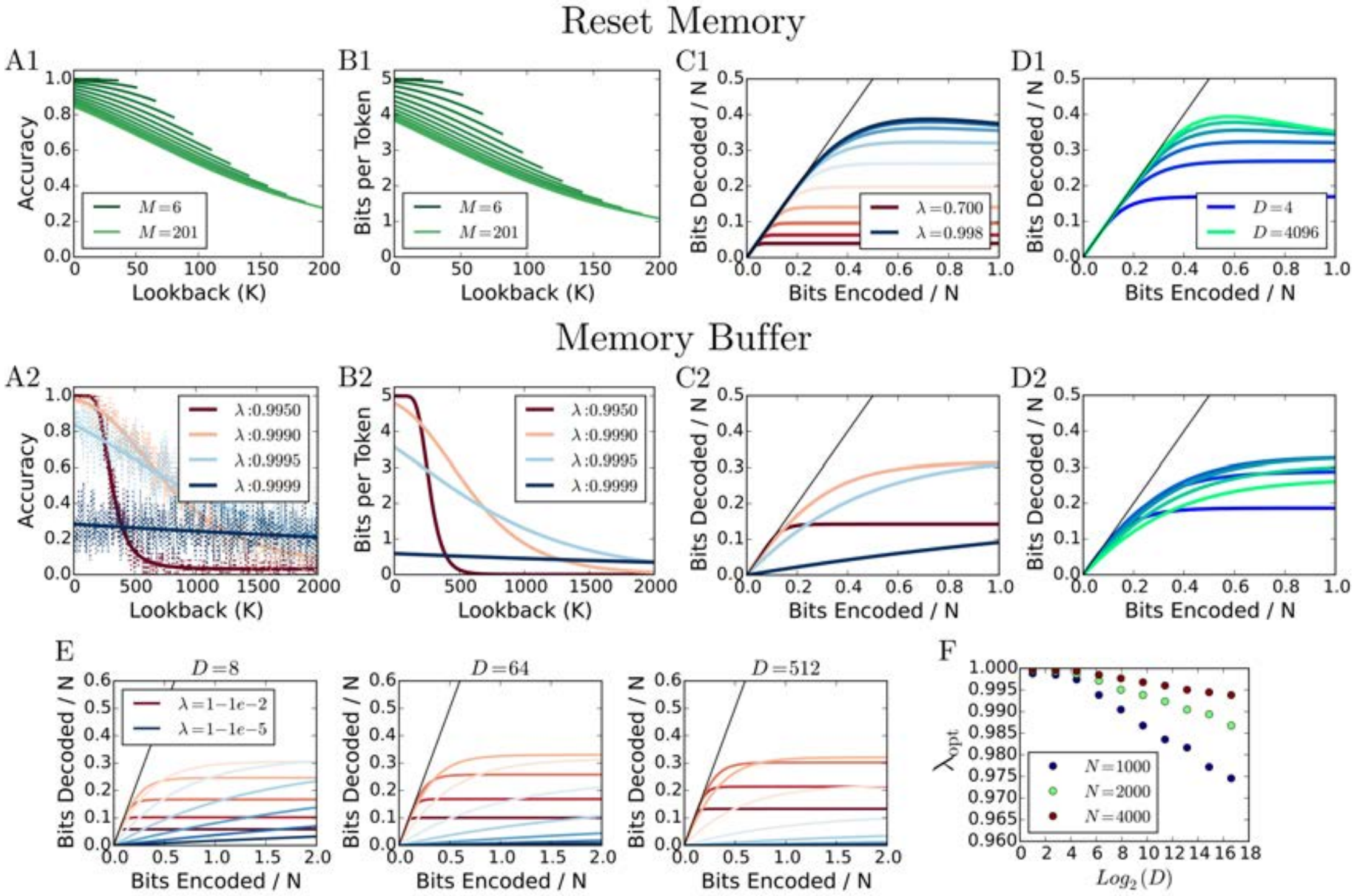

Figure 6: **Linear network with contracting recurrent weights.** A1. Accuracy in networks with $\lambda < 1$. Multiple evaluations of $p_{corr}$ are shown as a function of $K$ for sequences of different lengths, $M$. ($\lambda = 0.996$, $N = 1000$). B1. The information per item $I_{item}$ also depends on $K$. C1. The total retrieved information per neuron for different $\lambda$. The maximum is reached as $\lambda$ approaches 1 when $M$ is finite ($D = 64$; $N = 1000$). D1. The retrieved information is maximized as $D$ grows large ($\lambda = 0.988$). A2. Accuracy in networks with $\lambda < 1$ as $M \to \infty$ ($N = 10,000$; $D = 32$). B2. Information per item. C2. Total information retrieved as a function of the total information stored for different $\lambda$. There is a $\lambda$ that maximizes the information content for a given $N$ and $D$ ($D = 64$). D2. Total information retrieved as a function of the total information stored for different $D$ ($\lambda = 0.999$). Retrieved information is optimized by a particular combination of $D$ and $\lambda$. E. The total retrieved information per neuron versus the information stored per neuron for different $D$ and $\lambda$ over a wide range. As $D$ increases the information is maximized by decreasing $\lambda$. F. Numerically determined $\lambda_{opt}$ values that maximize the information content of the network with $M \to \infty$ for different $N$ and $D$.

rons is limited by $\kappa$:

$$f_\kappa(x) = \begin{cases} -\kappa & x \leq -\kappa \\ x & -\kappa < x < \kappa \\ \kappa & x \geq \kappa \end{cases} \tag{43}$$

Clipping functions of this type with specific $\kappa$-values play a role in VSAs which constrain the activation of memory vectors, such as the *binary-spatter code* (Kanerva, 1996) or the *binary sparse-distributed code* (Rachkovskij, 2001).

We will analyze the HDC encoding scheme, a network with a bipolar random input matrix and the recurrent weights a permutation matrix. With this the components of $\mathbf{x}$ will always assume integer values, and, due to the clipping, the components be confined to $\{-\kappa, -\kappa+1, ..., \kappa\}$. As a consequence, $z_{d,i}$, defined as in (30), will also assume values limited to $\{-\kappa, ..., \kappa\}$. To compute $s$, we need to track the mean and variance of $z_{d,i}$. This requires iterating the Chapman-Kolmogorov equation (3). To do so, we introduce a vector $\mathbf{q}$ with $q_{\mathcal{J}(k)}(m) := p(z_{d,i}(m) = k)\ \forall k \in \{-\kappa, ..., \kappa\}$, which tracks the probability distribution of $z_{d,i}$. The probability of each of the integers from $\{-\kappa, ..., \kappa\}$ is enumerated in the $2\kappa + 1$ indices of the vector $\mathbf{q}$, and $\mathcal{J}(k) = k + \kappa$ is a bijective map from the values of $z_{d,i}$ to the indices of $\mathbf{q}$ with inverse $\mathcal{K} = \mathcal{J}^{-1}$. To compute the sensitivity of a particular recall, we need to track the distribution of $z_{d,i}$ with $\mathbf{q}$ before the item of interest is added, when the item of interest is added, and in the time steps after storing the item of interest. Note that $\kappa$ is defined relative to the standard deviation of the codebook, $\kappa = \kappa^*/\sqrt{V_{\mathbf{\Phi}}}$. A simple scaling can generalize the following analysis to account for codebooks with different variance.

**Reset memory:**
At initialization $\mathbf{x}(0) = 0$, and so $q_j(0) = \delta_{\mathcal{K}(j)=0}$. For each time step that an input arrives in the sequence prior to the input of interest, a $+1$ or $-1$ will randomly add to $z_{d,i}$ up until the bounds induced by $f_\kappa$, and this can the be tracked with the following diffusion of $\mathbf{q}$:

$$q_j\,(m+1) = \frac{1}{2}\begin{cases} q_j(m) + q_{j+1}(m) & \text{when } \mathcal{K}(j) = -\kappa \\ q_{j-1}(m) + q_j(m) & \text{when } \mathcal{K}(j) = \kappa \\ q_{j-1}(m) + q_{j+1}(m) & \text{otherwise.} \end{cases} \quad \forall\, m \neq M - K \tag{44}$$

Once the vector of interest arrives at $m = M - K$, then all entries in $z_{d,i}$ will have $+1$ added. This causes the probability distribution to skew:

$$q'_j\,(m+1) = \begin{cases} 0 & \text{when } \mathcal{K}(j) = -\kappa \\ q_j(m) + q_{j-1}(m) & \text{when } \mathcal{K}(j) = \kappa \\ q_{j-1}(m) & \text{otherwise.} \end{cases} \quad m = M - K \tag{45}$$

The $K - 1$ inputs following the input of interest, will then again cause the probability distribution to diffuse further based on (44). Finally, $s(K)$ can be computed for this readout operation by calculating the mean and variance with $\mathbf{q}(M)$:

$$\mu(z_{d,i}) = \delta_{d=d'} \sum_{j=0}^{2\kappa} \mathcal{K}(j) q_j(M) \tag{46}$$

$$\sigma^2(z_{d,i}) = \sum_{j=0}^{2\kappa} (\mathcal{K}(j) - \mu(z_{d,i}))^2 q_j(M) \tag{47}$$

**Memory buffer:**
For $M \to \infty$ the diffusion equation (44) will reach a uniform equilibrium distribution, with the values of $z_{d,i}$ uniformly distributed between $\{-\kappa, ..., \kappa\}$: $q_j(\infty) = 1/(2\kappa + 1)\ \forall j$. This means, like with contracting recurrent weights, the clipped-linear function bounds the noise variance of the saturated state. Here the variance bound of $z_{d,i}$ is the variance of the uniform distribution, $((2\kappa+1)^2 - 1)/12$. Thus, information can still be stored in the network even after being exposed to an infinite sequence of inputs. The sensitivity in the saturated state can be calculated with $M \to \infty$ by replacing in (45) $\mathbf{q}(m)$ with $\mathbf{q}(\infty)$, and then again using the diffusion equation (44) for the $K - 1$ items following the item of interest.

Figure 7 illustrates this analysis of the distribution of $z_{d,i}$. When the item of interest is added, the probability distribution is most skewed and the signal degradation is relatively small. As more items are added later, the distribution diffuses to the uniform equilibrium, and the signal decays to 0. The figure compares operation with the initial states corresponding to reset memory and memory buffer: empty (Fig. 7 A1-F1) and saturated (Fig. 7 A2-F2). Numerical optimization of memory capacity shows how $\kappa_{opt}$ increases with $N$ and decreases with $D$, the parameter when $M \to \infty$ (Fig. 7 G).

### 2.4.3 Neurons with squashing nonlinear transfer function

The case when the neural transfer function $f(x)$ is a saturating or squashing function with $|f(x)|$ bounded by a constant also implies $z_{d,i}$ is bounded with $|z_{d,i}| \leq z_{max}$. For any finite fixed error, one can choose an $n$ large enough so that the distribution $p(z_{d,i} = k) = q_{\mathcal{J}(k)}$ can be approximated by discretizing the state space into $2n + 1$ equal bins in $\mathbf{q}$. Similar as for the clipped-linear transfer function, one can construct a bijecteve map from values to indices and track $\mathbf{q}$ approximately using rounding to discretize the distribution, $\mathcal{J}(k) = \lfloor \frac{n}{z_{max}}(k + z_{max}) \rceil$, with inverse $\mathcal{K} = \mathcal{J}^{-1}$. The Kolmogorov equation (3) simplifies in this discrete case and without neuronal noise to the following updates, one for encoding the signal, and one for encoding the distractors:

$$q_{j^*}(m+1) = \sum_{j=0}^{2n} \int dy\, p_{\mathbf{\Phi}}(y) q_j(m)\, \delta_{j^* = \mathcal{J}(f(\mathcal{K}(j)+y))} \quad \forall m \neq M - K \tag{48}$$

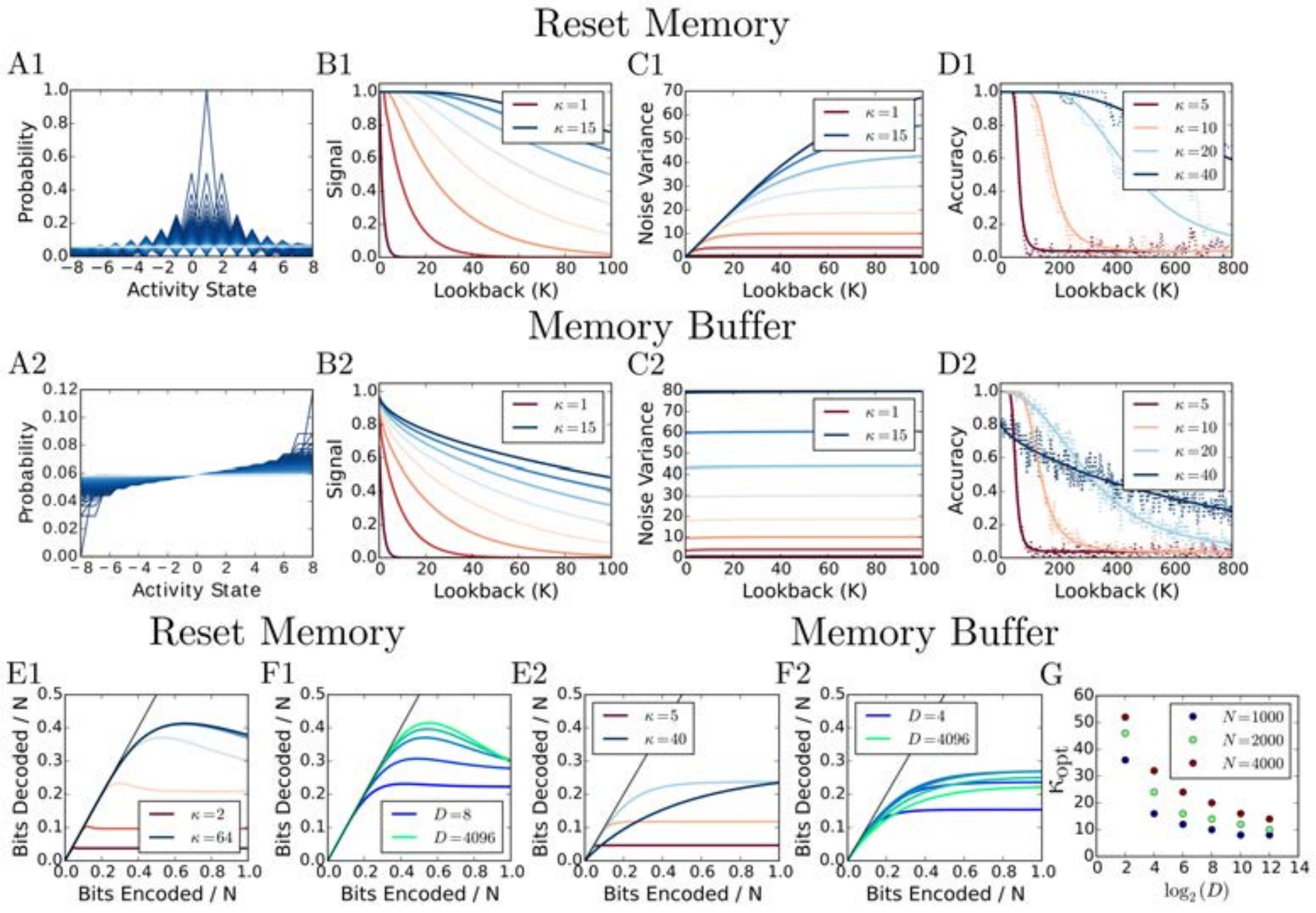

Figure 7: **Capacity for neurons with clipped-linear transfer function.** A1. The probability distribution of $z_{d,i}$ for retrieval of the first sequence item, as the sequence length is increased. The distribution evolves according to (44) and (45), it begins at a delta function (dark blue), and diffuses to the uniform equilibrium distribution when $M$ is large (light blue). B1. The clipped-linear function causes the signal to degrade as more items are stored ($M = K$; $N = 5000$; $D = 27$). C1. The variance of the distribution grows as more items are stored, but is bounded. D1. The accuracy theory fits empirical simulations decoding the first input as more input symbols are stored (dashed lines; $M = K$; $N = 5000$; $D = 27$). A2. The probability distribution of $\mathbf{z_{d,i}}$ for the memory buffer, that is, when $M \to \infty$. The most recent symbol encoded (dark blue) has the highest skew, and the distribution diffuses to the uniform equilibrium for readout further in the past (light blue). B2. The signal is degraded from crosstalk, and decays as a function of the lookback. C2. The noise variance is already saturated and stays nearly constant. D2. The accuracy exhibits a trade-off between fidelity and memory duration governed by $\kappa$. E1. With reset memory, the information that can be decoded from the network reaches a maximum when $\kappa$ is large ($D = 256$). F1. The capacity increases with $D$ ($\kappa = 20$). E2. When $M \to \infty$, there is a trade-off between fidelity and memory duration, a particular $\kappa$ value maximizes the retrieved information for a given $D$ and $N$ ($D = 256$). F2. For a given memory duration ($\kappa = 20$) an intermediate $D$ value maximizes the retrieved information. G. The memory duration $\kappa_{opt}$ that maximizes the retrieved information.

The update for the signal, given at $m = M - K$, where we know that $\mathbf{\Phi}_{d'}$ was stored in the network:

$$q'_{j^*}(m+1) = \sum_{j=0}^{2n} \int dy \; p_{\mathbf{\Phi}}(y) q_j(m) \; \delta_{j^*=\mathcal{J}(f(\mathcal{K}(j)+y^2))} \quad m = M - K \tag{49}$$

We illustrate our analysis for a network (1) with $\lambda = 1$ and $f(x) = \gamma \tanh(x/\gamma)$, where $\gamma$ is a gain parameter. As in the previous section, the network implements HDC coding, so the codebook is a bipolar i.i.d random matrix and the recurrent weights a permutation matrix. Our simulation experiments with this memory network examined both reset memories (Fig. 8 Row 1) and memory buffers (Fig. 8 Row 2). The iterative analysis (48, 49) can be used to compute the sensitivity and it predicts the experimentally observed readout accuracy very well. We find that a memory with the neural transfer function $\tanh$ possesses quite similar performance as a memory with the clipped-linear neural activation function.

### 2.4.4 Forgetting time constant and extensive capacity in memory buffers

We have analyzed different mechanisms of a recency effect in memory buffers, contracting weights and squashing, nonlinear neural activation functions. Here we will compare their properties and find parameters that optimize memory capacity.

For contracting weights, the *forgetting time constant* ($\tau$) is defined from the exponential decay of the sensitivity $\lambda^K$ in (42) by $\lambda = e^{-1/\tau}$ :

$$\tau(\lambda) = -1/\log \lambda \tag{50}$$

We derive the $N$ that optimizes the capacity of the memory buffer for a desired time constant (Fig. 9A).

The forgetting time constants for nonlinear activation functions can be computed by equating the bound of the variance induced by the nonlinearity to the bound induced by contracting weights. For clipping, the noise variance is bounded by the variance of the uniform distribution, $((2\kappa+1)^2 - 1)/12$, which can be equated to the bound of contracting weights $1/(1-\lambda^2)$. With (50) one obtains:

$$\tau(\kappa) = \frac{-2}{\log\left(1 - \frac{3}{\kappa(\kappa+1)}\right)} \approx \frac{2}{3}\kappa^2 \approx \frac{2}{3}\frac{(\kappa^*)^2}{V_{\mathbf{\Phi}}} \tag{51}$$

where the approximation holds for large $\kappa$.

Equating the bound of the noise variance to $1/(1-\lambda^2)$ is a general technique to approximate the time constant for any nonlinear function with (50). For the $\tanh$ nonlinearity we cannot compute analytically the forgetting time constant for a parameter value $\gamma$. Instead we use (48) and (49) to estimate its variance bound and equate it to $1/(1-\lambda^2)$.

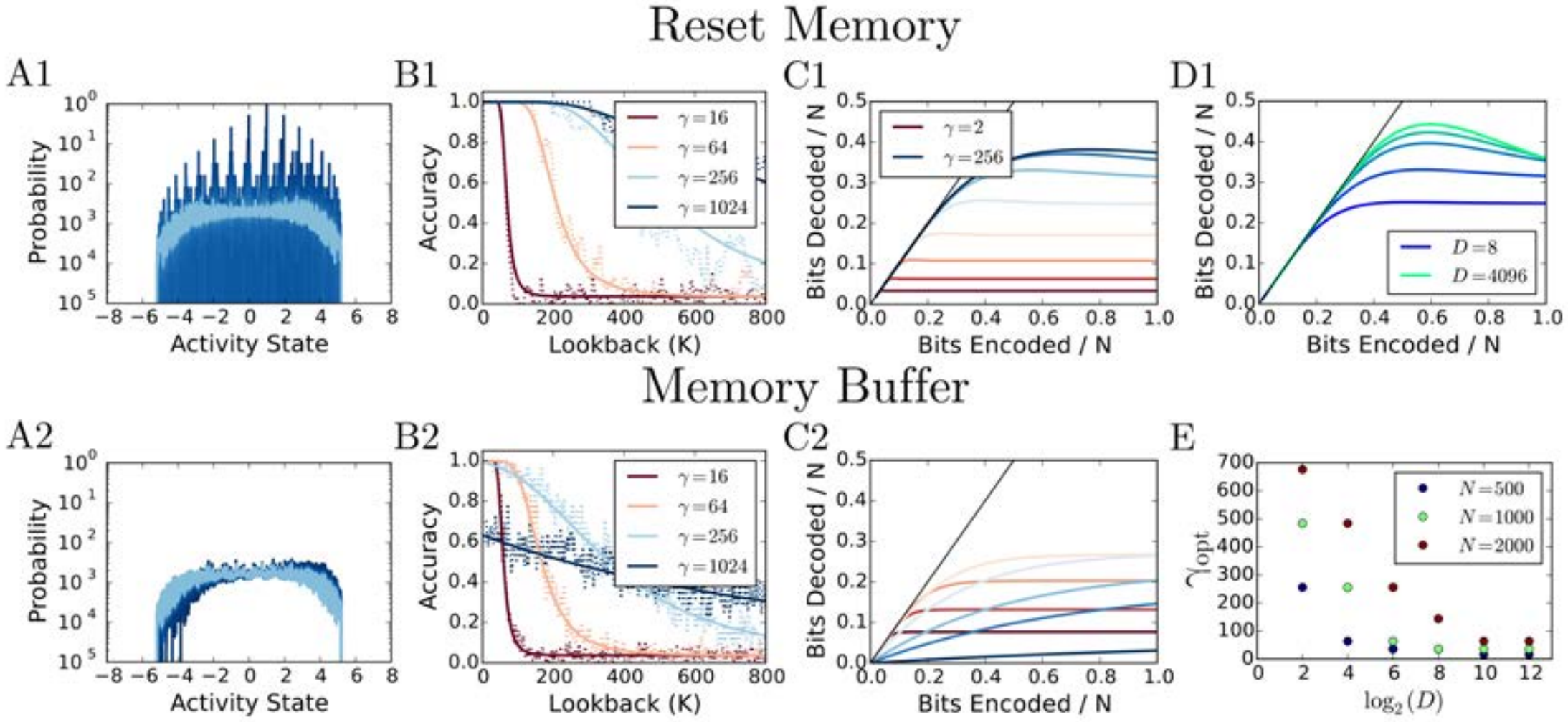

Figure 8: **Capacity for neurons with saturating nonlinearity** A1. The probability distribution of one term $z_{d,i}$ in the inner product used for retrieval of the first sequence item, as the sequence length is increased. The distribution begins as a delta function (dark blue) when only one input symbol is stored, and approaches the equilibrium distribution when $M$ is large (light blue). B1. The accuracy theory (solid lines) correctly describes simulations (dashed lines) retrieving the first input as more inputs are stored ($M = K$; $N = 2000$; $D = 32$). C1. Retrieved information as a function of the stored information. When $M$ is finite, the maximum is reached for large $\gamma$ ($D = 256$). D1. The capacity increases as $D$ increases ($\gamma = 64$). A2. The probability distribution of $\mathbf{z_{d,i}}$ when a new item is entered at full equilibrium, that is, when $M \to \infty$. The distribution for most recent input symbol possesses the highest skew (dark blue), and the distribution is closer to the uniform equilibrium (light blue) for input symbols encoded further back in the history. B2. The accuracy exhibits a trade-off between fidelity and memory duration governed by $\gamma$. C2. When $M$ is large, there is a $\gamma$ that maximizes the information content for a given $D$ and $N$ ($D = 256$). E. Numerically computed $\gamma_{opt}$ that maximizes the information content.

The relationships between $\tau$, the clipping parameter $\kappa$ (51), and the $\tanh$ gain parameter $\gamma$ (numerically estimated) are not far from linear in logarithmic coordinates (Fig. 9B). When the forgetting time constants are aligned, the accuracies of different recency mechanisms are not identical, but quite similar (Fig. 9C).

With neuronal noise, the optimal forgetting time constant decreases with noise variance (Fig. 9D), and also the memorized information (Fig. 9E). Interestingly, the memory capacity is independent of $N$ for a given noise variance (Fig. 9F), indicating that the memory buffer has extensive capacity. Since the effects of contracting weights and nonlinear neurons are similar, these networks can all achieve extensive memory in the presence of noise by keeping the time constant $\tau$ proportional to $N$.

With the clipped-linear activation function and HDC coding, for any finite setting of the clipping threshold $\kappa$, the number of bits required to represent or store the network state is finite and given by $I_{storage} = N \log_2(2\kappa + 1)$. One can now compute the ratio between the readout information and the information required to store the network state. In general, the amount of readout information increases with $\kappa$ (Fig. 7E1) and with $D$ (Fig. 7F1). However, as $\kappa$ increases so too do the number of bits required to represent the network. It turns out, there is an optimal ratio of about $0.08$, achieved with neurons that represent about 4-6 bits, the exact value dependent upon $D$ (Fig. 9G).

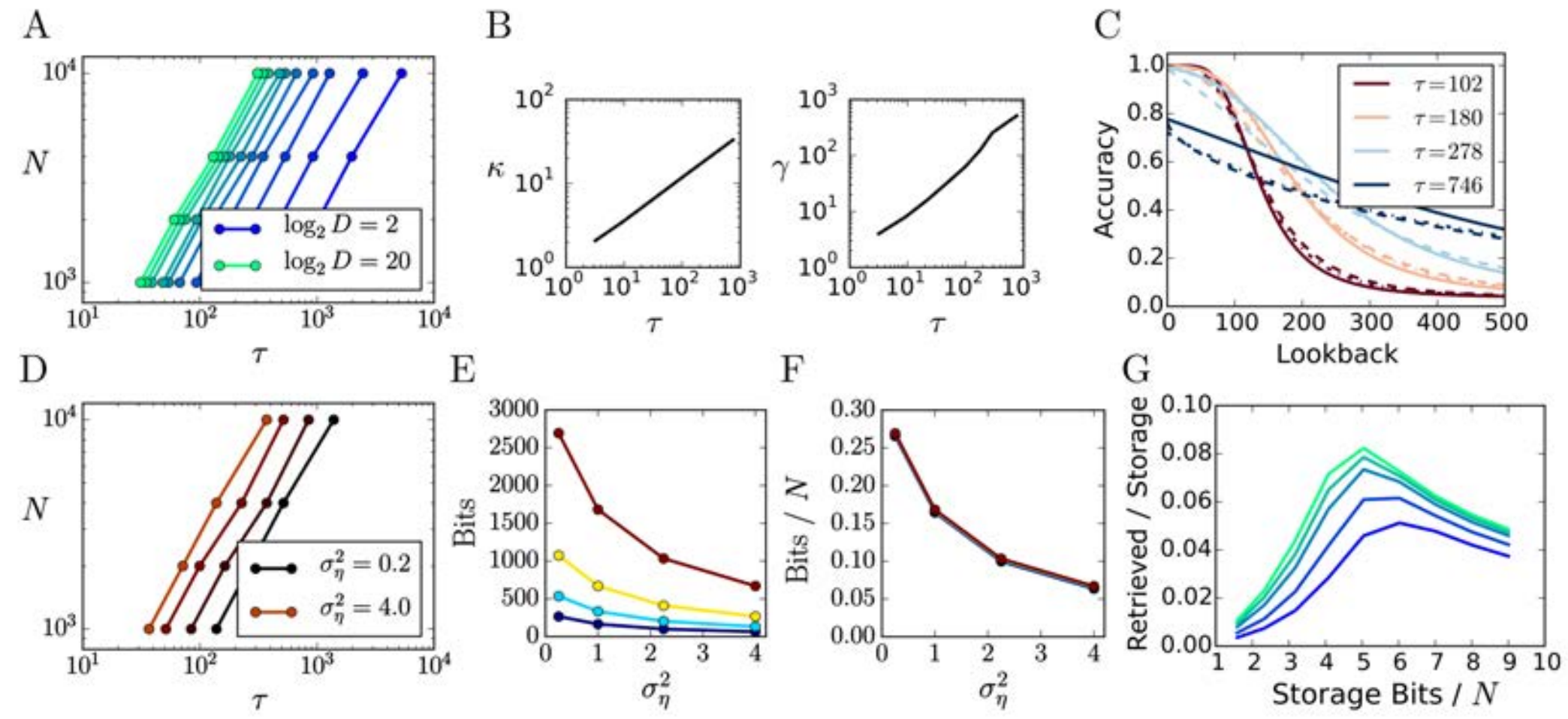

Figure 9: **Buffer time constant and optimization.** A. The optimal $N$ for given time constant $\tau$ and $D$. B. The relationship between time constant and nonlinear activation parameters, $\kappa$ and $\gamma$. C. Accuracy comparison of contracting (solid), clipped-linear (dashed) and $\tanh$ (dotted) networks which share the same time constant, found from the bound on noise variance. D. The optimal $N$ and $\tau$ shifts in the presence of noise ($D = 32$). E. The memory capacity decreases as noise increases ($N = [1K, 2K, 4K, 10K]$, blue to red). F. When capacity is normalized by number of neurons, the curves in panel E collapse to a single curve, showing capacity per neuron to be constant with $N$ and declining with neuronal noise. G. Ratio between retrieved and stored information for the clipped HDC network. The ratio is optimized with 4-5 bits of resolution per element ($D = [8, 32, 256, 1024, 4096]$, dark to light).

## 2.5 VSA indexing and readout for analog input vectors

The theory can be easily extended to the recall of coefficients of analog vectors. Rather than the input vector $\mathbf{a}(m)$ being a one-hot or zero vector, the input can be an arbitrary real vector and we wish to store and retrieve a sequence of such analog vectors in the network. We can derive information capacity under the assumption that the input vector is drawn independently from a normal distribution. In the following, the linear network with analog input is analyzed in two cases, operating as a reset memory (Results 2.5.1), and as a memory buffer (Results 2.5.2).

### 2.5.1 Capacity of reset memories with analog inputs

A sequence of vectors with analog coefficients $\mathbf{a}(m)$ is encoded into the network state by (1) with a random input matrix $\mathbf{\Phi}$ and unitary recurrent weight matrix $\mathbf{W}$. We return to considering reset memories with linear neurons, i.e., $f(\mathbf{x}) = \mathbf{x}$. During the encoding, each coefficient is indexed with a pseudo-random key vector. To readout an individual coefficient, we use in (2) a linear readout function $g(\mathbf{h}) = \mathbf{h}$, and the readout matrix of VSA models: $\mathbf{V}(K) = c^{-1}\langle \mathbf{a}(M-K)\mathbf{x}(M)^\top \rangle = c^{-1}\mathbf{W}^K\mathbf{\Phi}$, with $c = NE_{\mathbf{\Phi}}(x^2)$. The readout variable can be decomposed into $N$ contributions like in (30): $\hat{a}_{d'} = h_{d'} = \sum_i^N c^{-1} z_{d',i}$. For large enough $N$, $h_{d'}$ is distributed like a Gaussian due to the central limit theorem. We can compute $z_{d',i}$ from (30):

$$\begin{aligned} z_{d',i} &= (\mathbf{\Phi}_{d'})_i(\mathbf{W}^{-K}\mathbf{x}(M))_i \\ &= (\mathbf{\Phi}_{d'})_i\left[(\mathbf{\Phi}_{d'})_i a_{d'}(M-K)\right] \\ &+ (\mathbf{\Phi}_{d'})_i\left[\sum_{d\neq d'}^{D}(\mathbf{\Phi}_d)_i a_d(M-K) + \sum_{m\neq(M-K)}^{M}\left(\mathbf{W}^{M-K-m}\left(\sum_d^D \mathbf{\Phi}_d a_d(m)\right)\right)_i\right] \end{aligned} \tag{52}$$

The signal and the noise term are split onto two lines. In the expression $c^{-1}z_{d',i}$, the variance of the signal term is unity, and the resulting SNR is:

$$\begin{aligned} r = \frac{\sigma^2(a_{d'})}{\sigma^2(n_{d'})} &= \frac{N\sigma^2(a_{d'})}{\left(\sum_{d\neq d'} a_d^2(M-K) + \sum_{m\neq(M-K)}\sum_d a_d^2(m)\right)} \\ &= \frac{N}{(MD-1)} \approx \frac{N}{MD} \end{aligned} \tag{53}$$

When neuronal noise is present, the SNR becomes:

$$r = \frac{N}{MD}\left(\frac{1}{1+\sigma_\eta^2/(DV_{\mathbf{\Phi}})}\right) \tag{54}$$

If the input coefficients are all independent Gaussian random variables, then the total information can be computed with (28):

$$\frac{I_{total}}{N} = \frac{MD}{2N}\log_2(r+1) = \frac{\log_2(r+1)}{2r}\left(\frac{1}{1+\sigma_\eta^2/(DV_{\mathbf{\Phi}})}\right) \tag{55}$$

Note that the memory capacity for analog input is a function of $r$. Thus, the memory capacity is extensive when $MD$ is proportional to $N$. Without neuronal noise, the memory capacity depends on the product $MD$ and increasing either the sequence length or the alphabet size has the same effect on memory capacity.

We evaluated (55) in a wide parameter regime settings numerically optimized the memory capacity. We find linear extensive capacity (Fig. 10 A1). Interestingly, unlike in the symbolic case, there is no catastrophic forgetting – the retrieved information content saturates to a maximum value as the sequence length $M$ increases to infinity (Fig. 10B1). This means that in the limit of infinite sequence length, the information added by new data is perfectly cancelled by the information lost due to crosstalk (Fig. 10C1). The memory capacity is maximized for any large (finite) $M$ or $D$ compared to $N$, reaching $I_{total}/N = 1/(2\log 2)$. This capacity bound can be easily derived analytically, since it is achieved for $r \to 0$:

$$\begin{aligned}\frac{I_{total}}{N} &= \frac{\log(r+1)}{2r\log 2}\left(\frac{1}{1+\sigma_\eta^2/(DV_{\mathbf{\Phi}})}\right)\\ &=_{r\to 0} \frac{1}{2\log 2(1+\sigma_\eta^2/(DV_{\mathbf{\Phi}}))} \approx \frac{0.72}{1+\sigma_\eta^2/(DV_{\mathbf{\Phi}})}\end{aligned} \tag{56}$$

However, the regime of optimal memory capacity with $r \to 0$ is not interesting for applications. The critical question is, what fraction of this capacity is available for a fixed desired level of SNR $r^*$. The answer is depicted in Fig. 10D1, as a single curve. Because the memory capacity in the absence of neuronal noise depends only on the SNR, the curve describes all settings of the parameters $N$, $M$, and $D$. The capacity starts at the limit value (56) and then decreases as $r^*$ increases (Fig. 10E1). For instance, if the desired SNR is $r^* = 1$, one needs exactly as many neurons ($N$) as there are coefficients contained in the data sequence ($MD$), and achieves a memory capacity of 0.5 bits per neuron.

### 2.5.2 Capacity for memory buffer with analog input

Here, we analyze the memory buffer with linear neurons and contracting weights $\lambda < 1$, and an infinite sequence of analog input vectors, $M \to \infty$. To compensate for the signal decay, the linear readout contains a factor $\lambda^{-K}$:

$$\hat{\mathbf{a}}(M-K) = \mathbf{V}(K)^\top \mathbf{x}(M) = c^{-1}\lambda^{-K}\mathbf{\Phi}^\top\mathbf{W}^{-K}\mathbf{x}(M)$$

The readout produces then the original input, corrupted by Gaussian noise, $\hat{a}_{d'} = a_{d'} + n_{d'}$, with the SNR:

$$r(K) = \lambda^{2K}\frac{N(1-\lambda^2)}{D(1-\lambda^{2M})}\left(\frac{1}{1+\sigma_\eta^2/(DV_{\mathbf{\Phi}})}\right) \tag{57}$$

The similarity between memory buffers with different mechanisms of forgetting still holds for the case of analog input. For buffers with nonlinear neurons, one can use the analysis of the forgetting time constants in section 2.4.4 and use equation (50) to determine the corresponding value of $\lambda$. These values can be used in equation (57) to compute the SNR of the readout.

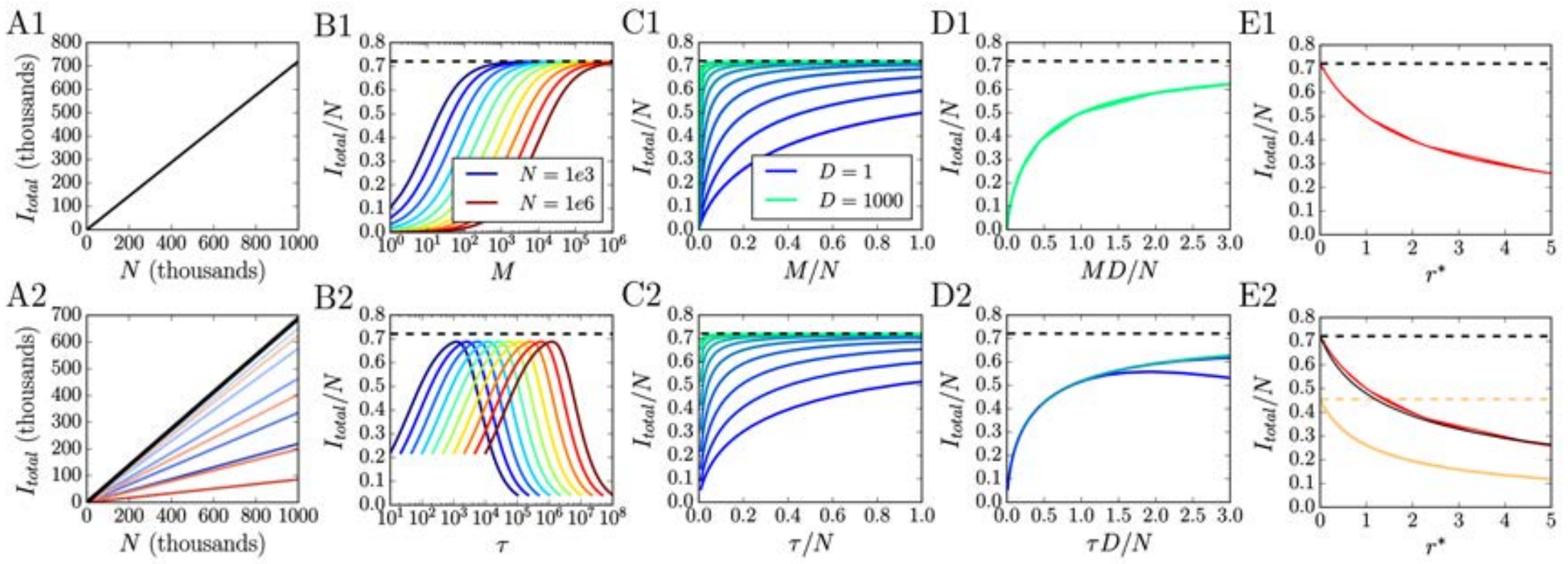

Figure 10: **Memory capacity of Gausian inputs.** A1. Numeric evaluation shows extensive memory for finite Gaussian input sequences. B1. As $M$ grows, the memory capacity saturates to the maximum. Note, each curve for different $N$ and fixed $D = 10$ is similar. C1. The family of curves in panel B1 reduces to a single curve with $M/N$ for each $D$. As $D$ grows the memory capacity grows to the bound for any $M$. D1. The family of curves in C1 reduces to a single curve with $MD/N$, which is directly related to $1/r$. Large $MD$ (small $r$) reaches the memory capacity limit. E1. The same function in D1 with $r$ as the x-axis. A2. Numeric evaluation shows extensive capacity whenever $\tau/N$ is held constant (colored lines), but a particular ratio results in the maximum (black line). B2. Similar curves for different $N$ and fixed $D = 10$ show an ideal $\tau$ that maximizes memory capacity. C2. The curves in B2 reduce to a single curve for each $D$ with the ratio $\tau/N$. As $D$ grows, the capacity is maximized. D2. As $\tau$ and $D$ grow large, the capacity saturates to the maximum. E2. The information per neuron retained at high-fidelity $I^*$ (copper) and the total mutual information per nueron (black) declines as the desired minimum $r^*$ increases. The total mutual information of the memory buffer behaves similarly to the total information in the linear case (compare black line to red line; red line same as in panel D1).

If the input is independent Gaussian random variables, the total information is:

$$I_{total} = \frac{D}{2} \sum_{K}^{M} \log_2 (r(K) + 1) \tag{58}$$

Inserting (57) into (58) we obtain:

$$I_{total} = \frac{D}{2} \log_2 ((-b_M q; q)_M) \tag{59}$$

where $q = \lambda^2$, $b_M := \frac{N(1-q)}{D(1+\sigma_\eta^2/(DV_\Phi))(1-q^M)}$ and $(a;q)_M$ is the $q$-Pochhammer symbol (see Methods 4.3.1). The advantage of formulation (59) is that it is well defined and can be properly used for $M \to \infty$.

If one numerically optimizes the memory capacity using (58) for $MD >> N$, like in the case of reset memories, one finds a linear extensive capacity (Fig. 10A2). Extensive capacity is retained for any constant ratio $\tau/N$, but there is a particular ratio that is the maximum (Fig. 10B2). A single curve for each value of $D$ describes the memory capacity as a function of the ratio $\tau/N$ (Fig. 10C2). As $D$ grows large and $\tau$ is optimized, the capacity saturates at the same asymptotic value as (56). By rescaling the x-axis with $D$, curves for different $D$ become very similar (Fig. 10D2), but not identical; if $\tau$ is too large for a fixed $D$ then the information starts to decrease. These curves collapsed to the exact same curve in the reset memory (Fig. 10D1), because the effects of $M$ and $D$ are fully interchangeable. However, the forgetting time constant introduces a distinction between the $M$ and $D$ parameters. The asymptotic memory capacity for the memory buffer (59) yields the same numeric value as for the reset memory, when $\tau$ is optimized:

$$\frac{I_{total}}{N} = \frac{1}{2\log(2)(1+\sigma_\eta^2/(DV_{\mathbf{\Phi}}))} \approx \frac{0.72}{1+\sigma_\eta^2/(DV_{\mathbf{\Phi}})} \tag{60}$$

This bound is assumed whenever $\tau$, $D$ both become large enough (see Methods 4.3.1).

Again the maximum memory capacity (60) is not relevant for applications, because the recall SNR for most memories is extremely low. To determine the usable capacity, we estimate $I^*(r^*) = \sum_{\{K:r(K)\geq r^*\}} I_{item}(K)$, the maximum information about past inputs that can be recalled with a given SNR $r^*$ or better. The optimum is reached when as many inputs as possible can be readout with SNR greater than $r^*$. From the condition $r(K^*) = r^*$ and equation (57) one finds the optimal time constant (Methods 4.3.2):

$$\frac{\tau_{opt}}{N} = \frac{2}{eDr^*(1+\sigma_\eta^2/(DV_{\mathbf{\Phi}}))} \tag{61}$$

The usable memory capacity for a given SNR threshold $r^*$ is plotted as the copper line in Fig. 10E2. Interestingly, the intercept of this curve for $r^* \to 0$ is lower than (60), the numeric capacity value can be computed (Methods 4.3.2) as:

$$\frac{I^*(r^* \to 0)}{N} = \frac{1-e^{-1}}{2\log(2)(1+\sigma_\eta^2/(DV_{\mathbf{\Phi}}))} \approx \frac{0.46}{1+\sigma_\eta^2/(DV_{\mathbf{\Phi}})} \tag{62}$$

The difference between the total capacity (60) and this result is the unusable fraction of information with $r(K) < r^*$.

If one counts usable and unusable information towards $I_{total}$ in a network which is optimized for $I^*$, another interesting phenomenon occurs: $I_{total}$ for a particular optimized $r^*$ of the buffer memory is very similar to the capacity of memory with reset (Fig. 10E2, black line, red line for comparison is same as panel E1). In both cases, the information capacity drops very similarly as the required fidelity $r^*$ is increased. Note, that the meaning of $r^*$ is different in both cases: with reset, it denotes the SNR for all memories; for the buffer, it is the lowest accepted SNR. Although the total capacity is so similar for reset memory and buffer, the usable information is different. With a reset, all information is retrieved with exactly fidelity $r(K) = r^*$. For the memory buffer, only the fraction depicted by the copper curve has fidelity $r(K) \geq r^*$, while inputs further back in the history (beyond the critical value $r(K^*) = r^*$) still take up a significant fraction of memory, but do not count towards $I^*$. The buffer has lower capacity because its exponential decay of the input is only an imperfect substitute of a reset – leaving behind a sediment of unusable information.

## 2.6 Readout with the minimum mean square estimator

Thus far, we have analyzed readout mechanisms in (2) that were proposed in the VSA literature. In contrast, in the area of reservoir computing a different readout has been studied, with a readout matrix determined by minimizing the mean square error between stored and recalled data. The readout is $\hat{\mathbf{a}}(M-K) = \mathbf{V}(K)^\top \mathbf{x}(M)$, with $\mathbf{V}(K)$ the minimum mean square estimator (MMSE; i.e. linear regression) or Wiener filter, given by $\mathbf{V}(K) = \mathbf{C}^{-1}\mathbf{A}(K)$. Here, $\mathbf{A}(K) := \left\langle \mathbf{a}(M-K)\mathbf{x}(M)^\top \right\rangle_R \in I\!R^{N \times D}$ is the empirical covariance between input and memory state over $R$ training examples, and $\mathbf{C} := \left\langle \mathbf{x}(M)\mathbf{x}(M)^\top \right\rangle_R \in I\!R^{N \times N}$ is the covariance matrix of the memory state. In this section, we will investigate this readout method and compare its performance with traditional VSA readout methods described in previous sections.

The MMSE readout matrix can be determined by solving a regression problem for an ensemble of synthetic input sequences. It involves generating $R$ synthetic input sequences of length $M$ and by encoding them in a neural network with one particular choice of input and recurrent matrix. This yields $R$ copies of state vectors, each encoding one of the synthetic input sequences. The readout matrix is now determined by solving the regression problem between state vectors and synthetic input sequences. The particular choice of input and recurrent matrices does not significantly affect the following results as long as the VSA indexing assumptions (4)-(7) are followed. We show results for the input matrix being a Gaussian random matrix and the recurrent weights being a permutation matrix.

### 2.6.1 Reset memory with MMSE readout

**Symbolic inputs:**
The MMSE readout does indeed significantly increase the capacity of reset memories in the regime where $MD \lesssim N$ (Fig. 11A, B). However, as $D$ grows larger, the performance of the MMSE readout falls back to the performance of VSA models. The comparison of the performances for symbolic input sequences is shown in Figure 11A,B.

To find out whether training both the input and decoding matrix, $\mathbf{\Phi}$ and $\mathbf{V}(K)$, has any advantages, we investigated the optimization of these matrices with back-propagation-through-time. A network of $N = 500$ linear neurons with fixed orthogonal recurrent matrix is fed a sequence of random input symbols and trained to recall the $K$th item in the sequence history. The cross-entropy between the recalled distribution and the one-hot input distribution is the error function. During training, the winner-take-all function in (2) is replaced with softmax. The network is evaluated each time step as more and more input symbols are encoded. After $500$ inputs are presented, the network is reset to 0 (see Methods 4.4.1 for further details).

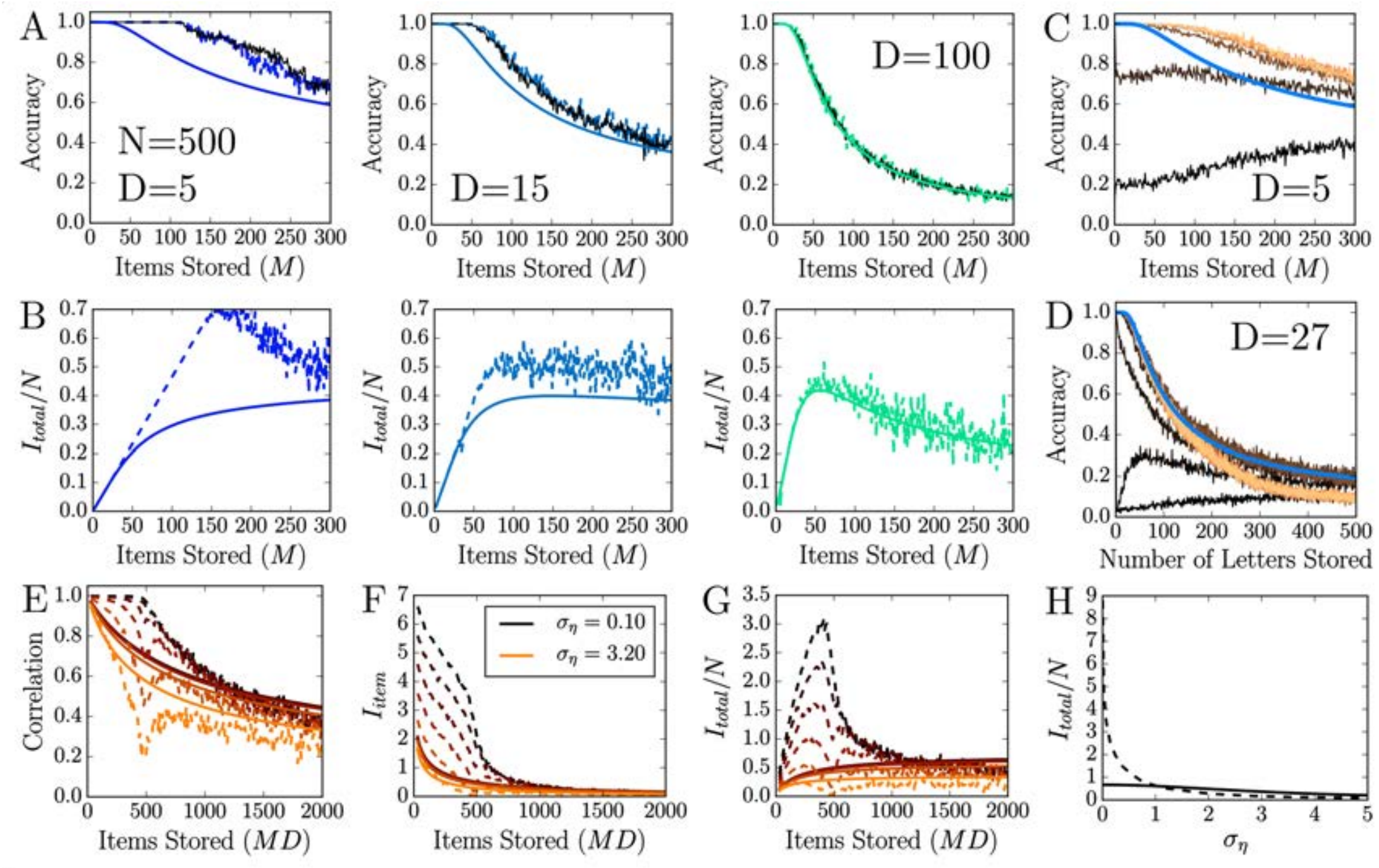

Figure 11: **MMSE readout of reset memory networks.** A, B. The accuracy (A) and capacity (B) of VSA prescribed readout (solid lines) is compared to MMSE readout (dashed lines) for different $D$ values. The empirical $s$ measured from the MMSE readout is plugged into $p_{corr}$ (12) and matches the measured accuracy (black lines). C, D. Accuracy of readout when both encoding and readout matrices are trained (black: early training iterations, copper: late training iterations, blue: VSA theory. After 500 inputs the network is reset.). Note, converged training in panel C (copper line) matches the dashed line in panel A. E-H. MMSE training for analog inputs with neuronal noise utilizes the regime where $MD \lesssim N$. ($N = 500$ in all panels).

The readout accuracy with backpropagation learning improves successively with training (black to copper in Fig. 11C,D), reaching the performance level of the MMSE readout (copper line in Fig. 11C matches dashed MMSE line in A). For larger $D$, the performance of backpropagation learning, MMSE readout, and VSA readout are equal (Fig. 11D). This convergence in performance shows that the simultaneous optimization of input and readout matrices (with backpropagation learning) yields no improvement over optimizing just the readout matrix for fixed input and recurrent matrix (with MMSE optimization).

**Analog inputs:**

Compared to the case with discrete input, for analog input the improvement achievable with MMSE has a similar pattern, but the magnitude of improvement is even more dramatic. When $MD \lesssim N$, then MMSE can greatly diminish the crosstalk noise and increase memory capacity (Fig. 11E-G). This improvement is because the MMSE readout allows the network to fully utilize all $N$ orthogonal degrees of freedom and store nearly $N$ analog numbers at high-fidelity. The retrievable information per symbol is increased to many bits (Fig. 11F), and the memory capacity is only limited by neuronal noise (Fig. 11H).

**Direct calculation of the readout matrix:**
The MMSE readout matrix for reset memory can also be computed without empirically solving the regression problem. The case where the number of synthetic input sequences is sent to infinity $R \to \infty$ (Methods 4.4.4) can be treated analytically. If $\mathbf{a}(m)$ has zero mean, the expected covariance matrix of the network states can be computed from $\mathbf{W}$ and $\mathbf{\Phi}$ as:

$$\tilde{\mathbf{C}} = \langle \mathbf{x}(M)\mathbf{x}(M)^\top \rangle_\infty = M\sigma_\eta^2 \mathbf{I} + \sum_{k=1}^{M} \mathbf{W}^k \mathbf{\Phi}\mathbf{\Phi}^\top \mathbf{W}^{-k} \tag{63}$$

An element of the expected covariance matrix is given by:

$$\tilde{C}_{ij} = (\delta_{i=j}) MDV_{\mathbf{\Phi}}(1 + \sigma_\eta^2/(DV_{\mathbf{\Phi}})) + (1 - \delta_{i=j}) \sum_{k=1}^{M} (\mathbf{W}^k)_i \mathbf{\Phi}\mathbf{\Phi}^\top (\mathbf{W}^{-k})_j \tag{64}$$

Further, the covariance between inputs and memory states converges to the VSA readout: $\tilde{\mathbf{A}}(K) = \langle \mathbf{a}(M-K)\mathbf{x}(M)^\top \rangle_\infty = \mathbf{W}^K \mathbf{\Phi}$. Thus, the MMSE readout matrix is given by:

$$\tilde{\mathbf{V}}(K) = \tilde{\mathbf{C}}^{-1} \mathbf{W}^K \mathbf{\Phi} \tag{65}$$

Note from (65) that the MMSE readout involves VSA readout with an additional multiplication by the inverse covariance of the memory state. Thus, dimensions in the state space that are only weakly driven through the input and recurrent matrices, are expanded to become fully useful in the decoding. The neuronal noise serves as ridge regularization of the regression, adding power in all dimensions. If the noise power $\sigma_\eta^2$ is positive, (65) is always well defined. However, without neuronal noise, the memory covariance matrix can become rank-deficient, and (65) undefined.

For symbolic input, this directly calculated linear filter matches the performance of the linear filter determined by linear regression from synthetic data. However for analog input, the filter determined by linear regression somewhat outperforms the directly calculated filter (see Methods 4.4.4; Fig. 19).

### 2.6.2 Memory buffer with MMSE readout

**Symbolic inputs:**

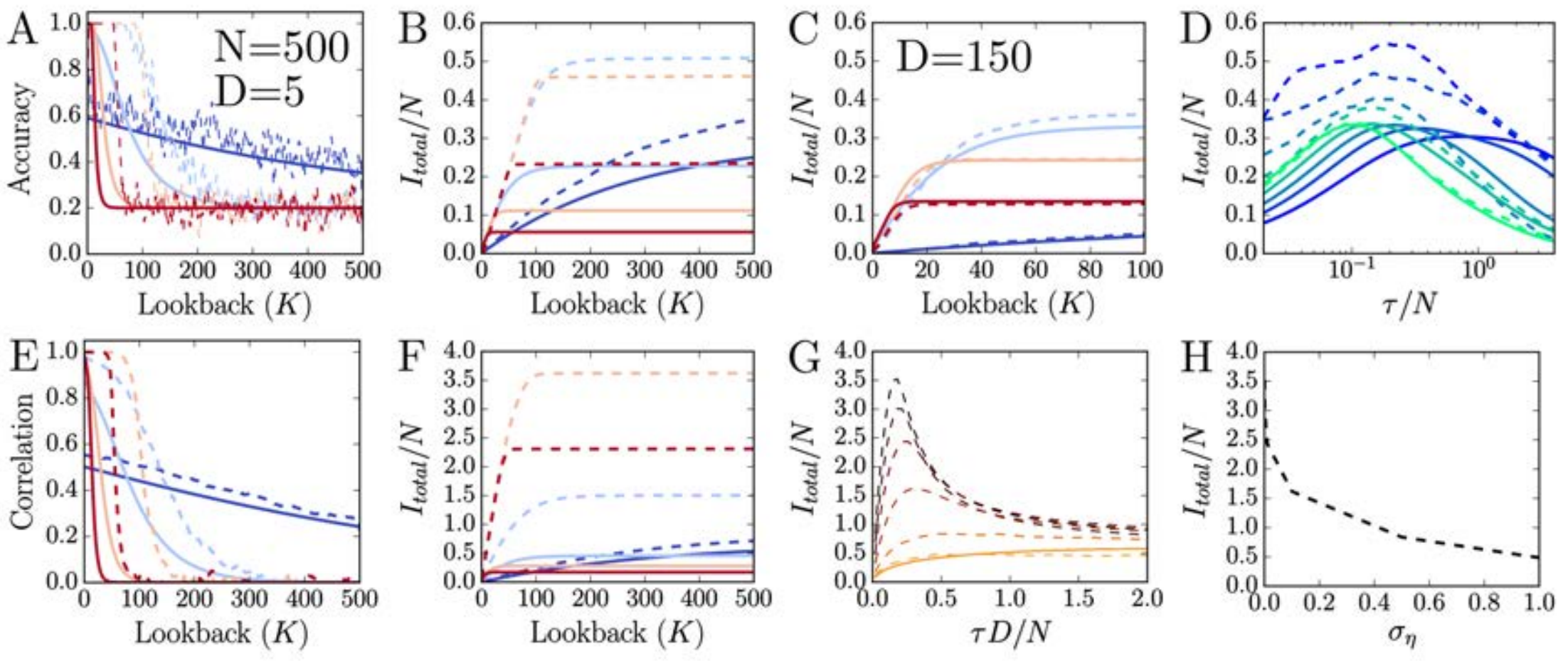

Figure 12: **MMSE readout in memory buffer networks.** A-C. The accuracy (A) and capacity (B,C) of MMSE readout (dashed lines) is compared to VSA readout (solid lines) for memory buffers with discrete inputs. Four networks with $\tau = 0.8, 0.95, 0.99, 0.999$. D. The capacity is computed as a function of the time constant for $D = 5, ..., 150$ (blue to green). For small $D$, the timeconstant can be used to enhance the readout accuracy and memory capacity, with optimal time constant between $(0.1 - 0.5)N$. E, F. The correlation (E) and capacity (F) for analog inputs shows that the MMSE training can utilize nearly all $N$ degrees of freedom with the right time constant ($N = 500$, $D = 5$). G, H. The capacity is computed for different amounts of noise.

Our findings for the memory buffer are not much different from our results for reset memories for both types of inputs. For discrete inputs and when $D$ is small, the MMSE readout improves the performance of memory buffers in the regime where $MD \lesssim N$, significantly increasing retrieval accuracy and capacity (Fig. 12A-D). Optimizing the time constant is still required to maximize memory capacity (Fig. 12D), which occurs when $\tau$ is $(0.1 - 0.5)N$.

**Analog inputs:**

For analog inputs, the time constant can be optimized to take advantage of the $N$ orthogonal degrees of freedom and mitigate crosstalk noise (Fig. 12E-G). High-fidelity retrieval can be maintained for $K_{max}$ items, with $K_{max} D \lesssim N$. Many bits per item can be recovered this way with $\tau$ between $(0.1 - 0.5)N/D$, and the memory capacity is again only limited by neuronal noise (Fig. 12H).

**Direct calculation of the readout matrix:**

The expected MMSE readout filter for the memory buffer is:

$$\tilde{\mathbf{V}}(K) = \tilde{\mathbf{C}}^{-1} \lambda^K \mathbf{W}^K \mathbf{\Phi} \tag{66}$$

For the case $D = 1$, this is the readout proposed in White et al. (2004).

The covariance matrix can be estimated for a given $\mathbf{W}$ and codebook $\mathbf{\Phi}$:

$$\tilde{\mathbf{C}} = \frac{\sigma_\eta^2}{1-\lambda^2}\mathbf{I} + \sum_{k=1}^{\infty} \lambda^{2k}\mathbf{W}^k\mathbf{\Phi}\mathbf{\Phi}^\top\mathbf{W}^{-k} \tag{67}$$

This can be simplified to a finite sum if the *cycle time* of the unitary matrix is known. Different unitary matrices can have different cycle times (random permutations have an expected cycle time of $0.62N$ (Golomb, 1964)), but we can pick a permutation matrix that has a maximum cycle time of $N$, s.t. $W^k = W^{k+N}$. This gives:

$$\tilde{\mathbf{C}} = \frac{\sigma_\eta^2}{1-\lambda^2}\mathbf{I} + \sum_{k=1}^{N} \frac{\lambda^{2k}}{1-\lambda^{2N}}\mathbf{W}^k\mathbf{\Phi}\mathbf{\Phi}^\top\mathbf{W}^{-k} \tag{68}$$

Here, the performance of MMSE with directly calculated readout matrix matches performance with the readout matrix obtained by regression over synthetic data.

## 2.7 Examples of storing and retrieving analog inputs

To illustrate the analog theory and also show that the input distribution can be complex and non-Gaussian, we encoded a random $(12\times12\times3)$ image patch into the vector $\mathbf{a}(m)$ each timestep. Several networks were created to act as memory buffers of the recently stored images of the same image sequence, with time constant held proportional to $N$, and we empirically evaluated their performance. The retrieved images are shown as a function of $N$ and lookback $K$ for networks with VSA readout (Fig. 13A), and networks with full MMSE readout (Fig. 13B). The empirically measured SNR of the MMSE readout is greatly increased compared to the VSA readout (Fig. 13C).

One advantage of VSAs is that they can be used to form arbitrary composite representations, and index data structures other than sequences. As a final example, we follow the procedure of Joshi et al. (2016) that uses the HDC code and algebra to encode the statistics of letter trigrams. They show how this technique can be used to create an effective, simple and low-cost classifier for identifying languages.

We performed the task of storing probabilities of letter $n$-grams by using the HDC algebra to create key vectors for each set of individual letters, bigrams or trigrams from the base set of $D = 27$ tokens (26 letters and space). The text of *Alice in Wonderland* served as the input to accumulate statistics about $n$-grams. Importantly to note here is that a complex combination of multiplication and permutation is needed to create the composite $n$-gram representation from the base set of random vectors. For instance, the trigram 'abc' is encoded as $\mathbf{x}_{abc} = \rho^2(\mathbf{\Phi}_a) \times \rho(\mathbf{\Phi}_b) \times \mathbf{\Phi}_c$, where $\rho$ is the permutation operation. This encoding distinguishes the 'abc' trigram from the trigram 'bac' or any other $n$-gram that may share letters or only differ in letter order.

Unlike the sequence indexing presented before, this composite representation cannot be created by a fixed matrix multiply. However, the composite binding and permutation operations still follow the statistics of assumptions (4)-(7), and thus are still characterized by our theory. In fact, all composite representations of discrete VSA base symbols ultimately follow these assumptions and are effectively a set of superposed random vectors. In terms of the neural network, the storage of $n$-gram statistics can be interpreted as the network accumulating the encoded $n$-gram vectors generated by external computations. The recurrent weight matrix is the identity, and thus the network counts up each $n$-gram vector that is given as input. The $n$-gram counts are indexed by the key vector made by the composite VSA operations and integrated into the memory state. We see that the empirical performance of such an encoding behaves in the same qualitative manner and matches the theory (Fig. 13D,E).

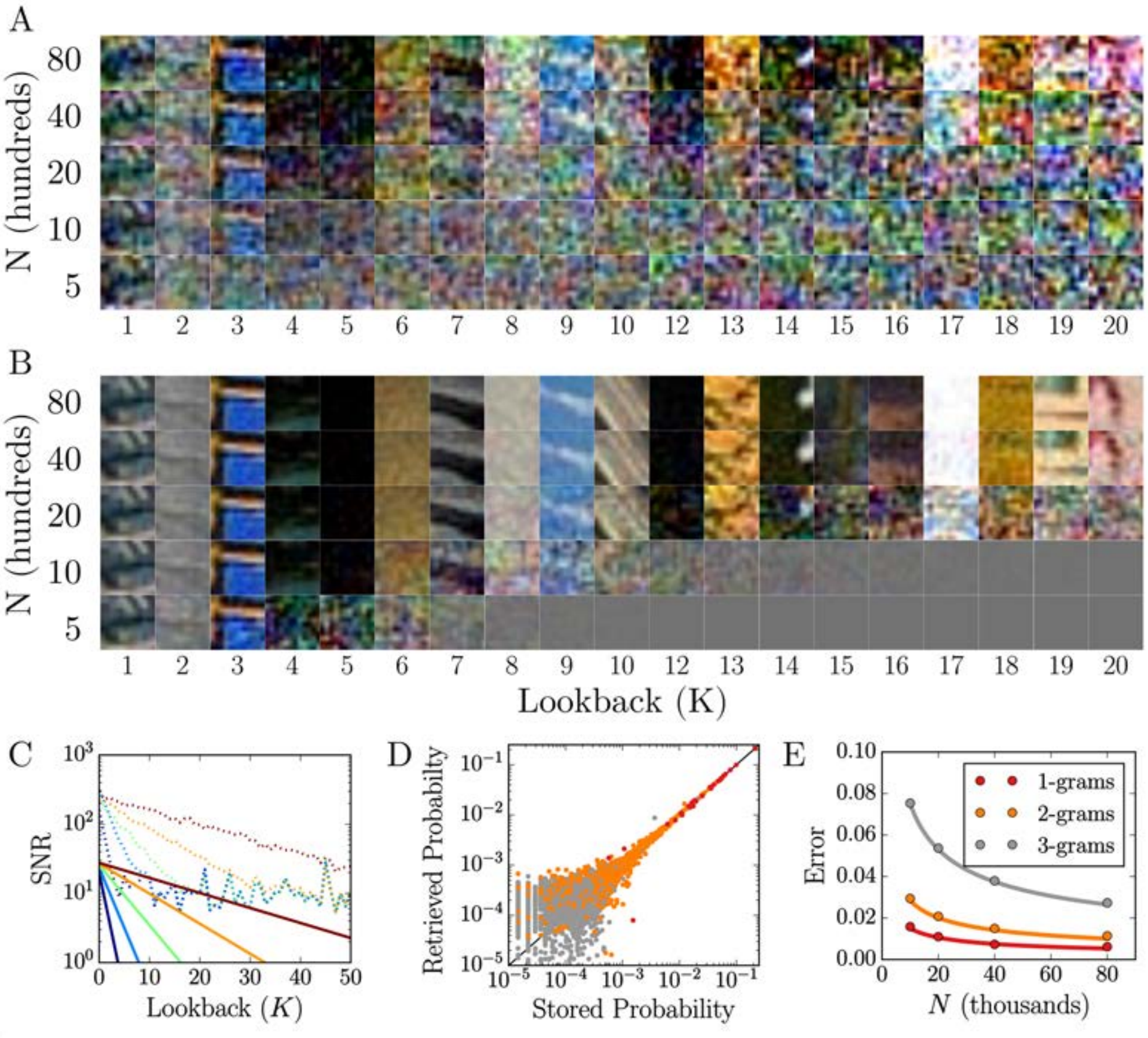

Figure 13: **Analog coefficient storage and retrieval.** A, B. A long sequence of image patches was stored in networks with different $N$ values, with $\tau$ proportional to $N$. The recent images were retrieved with VSA readout (A) and MMSE trained readout (B) and reconstructed for different lookback values, $K$. C. The measured SNR of MMSE readout (dashed lines) is greatly increased compared to the SNR of VSA readout. D. Indexing language $n$-grams with compositional binding operations has the same crosstalk properties as sequence indexing. E. The measured readout error of language statistics (dots) matches the theory (lines).

# 3 Discussion

The ability of recurrent neural networks to memorize past inputs in the current activity state is the basis of working memory in artificial neural systems for signal processing (Jaeger and Haas, 2004), cognitive reasoning (Eliasmith et al., 2012), and machine learning (Graves et al., 2014). What is often an intriguing but rather opaque property of recurrent neural networks, we try to dissect and understand in a systematic way. This is possible through the lens of vector-symbolic architectures (VSAs), a class of connectionist models proposed for structured computations with high-dimensional vector representations. Our work demonstrates that VSA sequence indexing can be mapped to recurrent neural networks with specific weight properties – the input weights random and the recurrent weights an orthogonal matrix. The computation of an iteration in such a network can now be concisely interpreted as generating a unique time stamp key and forming a key-value pair with the new input. These key-value bindings can be used to selectively retrieve data memorized by the network state. We were able to derive a theory describing the readout accuracy and information capacity achievable in such networks. Our results update and unify previous work on vector-based symbolic reasoning and contribute new results to the area of reservoir computing.

The theory includes two different forms of working memory. A reset memory operates like a tape recorder with start and stop buttons. The network state is initialized to zeros before input of interest arrives and the sequence of inputs is finite. With the ability to reset, networks without forgetting reach optimal capacity (Lim and Goldman, 2011). Existing VSA models map onto networks that operate as reset memories. In contrast, a memory buffer can track recent inputs in an infinite stream without requiring external reset. We investigated palimpsest networks as memory buffers, which attenuate older information gradually by various mechanisms, such as contracting recurrent weights or saturating neural nonlinearities. We showed that there is one essential property of palimpsest memories, the forgetting time constant. We demonstrate how this constant is computed for different forgetting mechanisms and that the model performances are very similar if the forgetting time constant is the same. Further, we demonstrated that the time constant can be optimized for obtaining high and extensive information capacity, even in the presence of neuronal noise.

The theory analyzes memory networks for two different types of input data, symbolic or analog. Symbolic inputs, encoded by one-hot and binary input vectors, corresponds to neural network models of vector-symbolic architectures (VSA) (Plate, 1994, 2003; Gallant and Okaywe, 2013). In these models, a naive linear readout is followed by a nonlinear winner-take-all operation for additional error correction. The naive linear readout consists in projecting the memory state in the direction of the key vector associated with the wanted input. In addition to storing sequences, VSAs also allow to store other composite structures, such as unconnected sets of key-value pairs, trees, stacks, $n$-grams, etc., and our theory also extends to those (Fig. 13D,E).

The case of analog inputs has been considered before in reservoir computing (Jaeger, 2002; White et al., 2004; Ganguli et al., 2008). Some of these models have weight properties that fall outside the conditions of our theory. However, the regime described by the theory is particularly interesting. Previous work has shown that unitary recurrent weights optimize capacity (White et al., 2004). The readout in memories for analog data is typically linear. Going beyond the naive readout used in VSAs, these models use the Wiener filter providing the MMSE regression solution (White et al., 2004). The MMSE readout greatly improves performance when encoding independent analog values, as nearly the full $N$ orthogonal degrees of freedom can be used with minimal crosstalk noise. In other regimes, however, the naive readout in VSAs has advantages, as it yields similar performance and is much easier to compute[2].

## 3.1 Working memory in vector-based models of cognitive reasoning

**Analysis of existing VSA models:**
We demonstrated how various VSA models from the literature can be mapped to equivalent reset memories with linear neurons and unitary recurrent matrix. This approach not only suggests concrete neural implementations of VSA models, but also allowed us to trace and highlight common properties of existing models and develop a common theory for analyzing them.

VSA models use the fact that random high-dimensional vectors are nearly orthogonal, and that the composition operations (such as forming a sequence) preserve independence and generate pseudo-random vectors. The prerequisites of VSAs are formalized by the conditions (4)-(7).

[2] Computing the inverse of the $\mathbb{R}^{N \times N}$ matrix $\mathbf{C}$ scales with $O(N^3)$ and becomes unfeasible with $N$ larger than a few thousand. For instance, the computation of $\mathbf{C}^{-1}$ took over 14 hours for the $N = 8,000$ network shown in Fig. 13B

The previous analyses of specific VSA models (Plate, 1994, 2003; Gallant and Okaywe, 2013) were limited to the high-fidelity regime. Our theory yields more accurate estimates in this regime, revealing that the working memory performance is actually superior than previously thought. Specifically, we derived a capacity bound of 0.39 bits per neuron in the high-fidelity regime.

Importantly, our theory is not limited to the high-fidelity regime but predicts the effects of crosstalk and neuronal noise in all parameter regimes. In particular, the theory revealed that different VSA frameworks from the literature have universal properties. For larger networks, recall accuracy is independent of the moments of a particular distribution of random codes and therefore, the sensitivity for recall is universally $s = \sqrt{N/M}$ (35). This finding explains that achieving large sensitivity for high-fidelity recall requires the number of memory items to be smaller than the number of neurons. These results can be used to design optimized VSA representations for particular hardware (Rahimi et al., 2017).

**New VSA models with optimal readout:**
VSA models use a linear readout which is suboptimal but fast to compute. Here, we asked how much optimal linear readout with the minimal least square estimator (MMSE) or Wiener filter can improve the performance of VSA models. VSA models are used to index both symbolic and analog input variables, and the input distribution has important consequences on the coding and capacity.

Symbols are represented by binary one-hot vectors, and have an entropy of $M \log_2 D$. The MMSE readout can reduce crosstalk when the input can be encoded with fewer than $N$ numbers. However, the one-hot vectors still require $MD$ numbers to encode the input sequence, and the MMSE loses its advantage if $D$ is large (Fig. 11A,B). Discrete input sequences which have $MD > N$ can still be stored with the randomized code vectors as the basis because the entropy is only $M \log_2 D < N$, but MMSE training does not much improve readout accuracy in this regime.

The analog input vectors we considered have independent entries. In this case, what matters for retrieval accuracy is the number $MD$ of real numbers to store. Thus, the dimensions of the input vector, $D$, and the sequence length, $M$, contribute in the same way to the memory load. If and only if $MD \lesssim N$, the MMSE readout can remove crosstalk noise inherent in standard VSA models, and can greatly increase the capacity to many bits per neuron. However, if $MD > N$, the performance of MMSE readout drops back to the performance of standard VSA readout.

**New VSA memory buffers:**
Plate (1994) describes *trajectory-association*, the mechanisms for indexing an input sequence with a RNN. Our analysis quantifies how recency effects, caused by contracting weights or nonlinear activations, influence recall accuracy. This analysis enabled us to construct memory buffers that are optimized to perform trajectory-association in continuous data streams. For instance, consider an optimized digital implementation of a dis-crete memory buffer. We can use HDC code framework, $p_{\mathbf{\Phi}}(x) = \mathcal{B}_{0.5}, x \in \{-1, +1\}$, with clipping nonlinearity, and create a memory buffer where neurons only have integer activation states bounded by $\kappa$. Thus, the memory state can be represented by $N \log_2(\kappa)$ bits (Fig. 9G). The recurrent matrix is a simple permutation. This is an efficient way to utilize the coding scheme for a digital device that can continually encode a spatio-temporal input into an indexed, distributed memory state. Readout neurons can be trained to recognize temporal input patterns (Kleyko et al., 2017). The complex vector representations used in FHRR could be useful for emerging computational hardware, such as optical or quantum computing.

## 3.2 Contributions to the field of reservoir computing

**Memory buffers have extensive capacity with optimized forgetting time constant**: Our theory captures working memory in neural networks that have contracting weights or saturating nonlinear neurons, both of which produce a recency effect. We derive for these diverse mechanisms the one single feature, which is critical to the memory performance, the network's forgetting time constant. If the forgetting time constant is the same for networks with different forgetting mechanisms, the memory performance becomes very similar (Figure 9). Putting the forgetting time constant in the center focus, enables a unified view on a large body of literature investigating the scaling and capacity of recurrent neural networks (Jaeger, 2002; White et al., 2004; Ganguli et al., 2008; Hermans and Schrauwen, 2010; Wallace et al., 2013).

Importantly, we found that memory buffers can possess extensive capacity in the presence of accumulating noise (Fig. 9E; enhanced by MMSE readout Fig. 12). To preserve extensive capacity with noise, the time constant ($\tau$) has to be chosen appropriately for given noise power ($\sigma_\eta^2$) and number of neurons ($N$). As $N$ grows large, the time constant of the network must also increase proportionally (Fig. 9A). With this choice, the noise accumulation scales but does not destroy the linear relationship between $N$ and the network capacity (Fig. 9D). It was first noted in White et al. (2004) that extensive capacity could be attained in normal networks (e.g. networks in Results 2.4.1) by correctly setting the decay parameter in relation to $N$.

Since the dynamic range of the neurons determines the time constant of the recency effect, it must be optimized with $N$ to achieve extensive capacity. Conversely, if the parameters related to the time constant are kept fixed as $N$ grows large, then the memory capacity does not scale proportionally to $N$. Thus, our result of extensive capacity is not in contradiction to (Ganguli et al., 2008) where a sub-extensive capacity of $O(\sqrt{N})$ is derived for a fixed dynamic range (e.g. $\kappa$) in networks with non-normal recurrent matrix. Our analysis confirms the sub-extensive capacity when the time constant is held fixed as $N$ grows. With normal matrices, we see $O(\log N)$ scaling with fixed dynamic range. This logarithmic scaling matches results reported in Wallace et al. (2013).

We show that extensive capacity can be achieved by increasing the dynamic range of the neurons as $N$ grows. Extensive capacity is achieved in this case, but only if the neuronal noise is fixed, and not growing proportionally to the dynamic range. If the noise assumptions are such that neuronal noise and dynamic range are proportional (Ganguli et al., 2008), then extensive capacity is not possible.

The time constant can also be increased by decreasing the variance of the codebook, $V_{\mathbf{\Phi}}$, and keeping the dynamic range fixed (51). In the zero-noise case, optimizing the codebook variance can achieve extensive capacity with a fixed dynamic range. However, when fixed noise is present, reducing the variance also increases the impact of noise (37), which prohibits extensive capacity.

The presented analysis of the recency effect has implications for other models which use principles of reservoir computing to learn and generate complex output sequences (Sussillo and Abbott, 2009). The buffer time constant $\tau$ and its relationship to network size could be used for optimizing and understanding network limits and operational time-scales.

Previous studies have suggested that the RNN (1) can memorize sparse input sequences in which the sequence length is greater than the number of neurons, $M > N$, if sparse inference is used in the readout (Ganguli and Sompolinsky, 2010). Using theory of *compressed sensing*, it has been shown for a fixed accuracy requirement that the number of required neurons scales as $N \propto Mp_s \log^{\ell}(M)$ (Charles et al., 2014, 2017). Our result for sparse input sequences, $N = s^2 Mp_s$, lacks the factor logarithmic with sequence length. One reason for this discrepancy is that our result requires the conditions (4)-(7) to hold. For large enough $M$, condition (7) will inevitably fail to hold because unitary matrix powers eventually loop, and thus the sequence indexing keys for long sparse input sequences are not independent. Thus, compressed sensing theory might account for this gradual failing of producing independent time stamp keys, which reduces performance as the sequence length grows. Another reason for disagreement is that the compressed sensing readout requires sparse inference of the complete sequence history – it does not permit access to individual sequence entries. In contrast, the readout procedures presented in this paper permit individual entries in the sequence to be retrieved separately.

**Optimal memory capacity of neural reservoirs:**
White et al. (2004) proposed *distributed shift register* models (DSR) that store a one-dimensional temporal input sequence "spatially" in a delay line using a connectivity that avoids any signal mixing. This model shows how the full entropy of an RNN can be utilized, and that capacity scales proportionally to $N$. The DSR is often cited as the upper bound of memory performance in reservoir computing. This is achieved by using a constructed code for indexing and storing sequence history along each of the $N$ neural dimensions, without any crosstalk. However, because of this construction, the network can break down catastrophically from small perturbations such as the removal of a neuron.

White et al. (2004) also analyze normal networks that can distribute the input sequence across the neurons and create a more robust memory. They used an annealed approximation for describing the readout performance for one-dimensional input sequences, which shares many characteristics with our assumptions (4)-(7). Their approximation includes the MMSE readout, and corresponds to analog memory buffer networks considered in Results 2.6.2. This theory suggests much lower performance of normal networks compared to the DSR. We directly compare the White et al. (2004) theory to memory buffers with naive VSA readout and MMSE readout in Methods 4.4.2.

We show how VSAs can be used to perform indexing with random code vectors for inputs of arbitrary dimensions. Our theory precisely characterizes the nature of crosstalk noise, and we show how MMSE readout can remove much of this crosstalk noise in the regime where $MD \lesssim N$. We see that VSA indexing with MMSE readout outperforms the White et al. (2004) theory for normal networks and can reach the memory performance of the DSR. Compared to the constructed codes like the DSR, there is a small reduction of the capacity due to duplication of code vectors of random codes (Methods 4.4.3), but this reduction becomes negligible for large network sizes. Thus, VSA encoding with MMSE readout can be distributed and robust, while still retaining the full capacity of the DSR.

## 3.3 Survey of memory capacity across models

Our results show that distributed representations indexed by VSA methods can be optimized for extensive capacity. We report the idealized, zero-noise memory capacity bounds found for different input streams and readouts in Table 1. The table reveals that the bounds are finite except for the case of MMSE readout with analog variables. The table summarizes the qualitative nature of different models, although these bounds may not be achievable under realistic conditions.

| Capacity $\left(\frac{bits}{neuron}\right)$ | VSA | | MMSE | |
|---|---|---|---|---|
| | reset | buffer | reset | buffer |
| symbolic | $\approx 0.5$ | $\approx 0.3$ | 1 | 1 |
| analog | 0.72 | 0.46 | $\infty$ | $\infty$ |

Table 1: **Idealized bounds on memory capacity.**

Table 2: Summary of VSA computing frameworks. Each framework has its own set of symbols and operations on them for addition, multiplication, and a measure of similarity.

| VSA | Symbol Set $(p_{\mathbf{\Phi}}(x))$ | Binding | Permutation | Trajectory Association |
|---|---|---|---|---|
| HDC | $\mathcal{B}_{0.5} : x \in \{-1, +1\}$ | $\times$ | $\rho$ | $\sum \rho^m(\mathbf{\Phi}_{d'})$ |
| HRR | $\mathcal{N}(0, 1/N)$ | $\circledast$ | none | $\sum \mathbf{w}^m \circledast \mathbf{\Phi}_{d'}$ |
| FHRR | $\mathcal{C} := e^{i\phi} : \phi \in \mathcal{U}(0, 2\pi)$ | $\times$ | $\circledast$ | $\sum \mathbf{w}^m \times \mathbf{\Phi}_{d'}$ or $\sum \mathbf{w}^{\circledast m} \circledast \mathbf{\Phi}_{d'}$ |
| MBAT | $\mathcal{N}(0, 1/N)$ | matrix | matrix | $\sum \mathbf{W}^m \mathbf{\Phi}_{d'}$ |

# 4 Methods

## 4.1 Vector symbolic architectures

### 4.1.1 Basics

The different vector symbolic architectures described here share many fundamental properties, but also have their unique flavors and potential advantages/disadvantages. Each framework utilizes random high-dimensional vectors (hypervectors) as the basis for representing symbols, but these vectors are drawn from different distributions. Further, different mechanisms are used to implement the key operations for vector computing: *superpostion*, *binding*, *permutation*, and *similarity*.

The *similarity* operation transforms two hypervectors into a scalar that represents similarity or distance. In HDC, HRR, FHRR and other frameworks, the similarity operation is the dot product of two hypervectors, while the Hamming distance is used in the frameworks which use only binary activations. The distance metric is inversely related to the similarity metric. When vectors are similar, then their dot product will be very high, or their Hamming distance will be close to 0. When vectors are orthogonal, then their dot product is near 0 or their Hamming distance is near 0.5.

When the *superposition* $(+)$ operation is applied to a pair of hypervectors, then the result is a new hypervector that is similar to each one of the original pair. Consider HDC, given two hypervectors, $\mathbf{\Phi}_A$, $\mathbf{\Phi}_B$, which are independently chosen from $p_{\mathbf{\Phi}}(x) = \mathcal{B}_{0.5} : x \in \{-1, +1\}$ and thus have low similarity ($\mathbf{\Phi}_A^\top \mathbf{\Phi}_B = 0 + noise$), then the superposition of these vectors, $\mathbf{x} := \mathbf{\Phi}_A + \mathbf{\Phi}_B$, has high similarity to each of the original hypervectors (e.g. $\mathbf{\Phi}_A^\top \mathbf{x} = N + noise$). In the linear VSA frameworks (Kanerva, 2009), Plate (2003), we do not constrain the *superposition* operation to restrict the elements of the resulting vector to $\{-1, +1\}$, but we allow any rational value. However, other frameworks (Kanerva, 1996; Rachkovskij and Kussul, 2001) use clipping or majority-rule to constrain the activations, typically to binary values.

The *binding* operation ($\times$) combines two hypervectors into a third hypervector ($\mathbf{x} := \mathbf{\Phi}_A \times \mathbf{\Phi}_B$) that has low similarity to the original pair (e.g. $\mathbf{\Phi}_A^\top \mathbf{x} = 0 + noise$) and also maintains its basic statistical properties (i.e. it looks like a vector chosen from $p_{\mathbf{\Phi}}(x)$). In the HDC framework, the hypervectors are their own multiplicative self-inverses (e.g. $\mathbf{\Phi}_A \times \mathbf{\Phi}_A = \mathbf{1}$, where $\mathbf{1}$ is the binding identity), which means they can be "dereferenced" from the bound-pair by the same operation (e.g. $\mathbf{\Phi}_A \times \mathbf{x} = \mathbf{\Phi}_B + noise$). In the binary frameworks, the binding operation is element-wise XOR, while in HRR and other frameworks binding is implemented by circular convolution ($\circledast$).

In different VSA frameworks, these compositional operations are implemented by different mechanisms. We note that all the binding operations can be mapped to a matrix multiply and the frameworks can be considered in the same neural network representation. The FHRR framework is the most generic of the VSAs and can utilize both multiply ($\times$) and circular convolution ($\circledast$) as a binding mechanism.

### 4.1.2 Implementation details

The experiments are all implemented in python as jupyter notebooks using standard packages, like numpy.

The experiments done with different VSA models use different implementations for binding, most of which can be captured by a matrix multiplication. However, for efficiency reasons, we implemented the permutation operation $\rho$ and the circular convolution operation $\circledast$ with more efficient algorithms than the matrix multiplication. The permutation operation can be implemented with $O(N)$ complexity, using a circular shifting function (`np.roll`). Efficient circular convolution can be performed by fast Fourier transform, element-wise multiply in the Fourier domain, and inverse fast Fourier transform, with $O(NlogN)$ complexity.

To implement FHRR, we utilized a network of dimension $N$, where the first $N/2$ elements of the network are the real part and the second $N/2$ elements are the imaginary part. Binding through complex multiplication is implemented as:

$$\mathbf{u} \times \mathbf{v} = \begin{bmatrix} \mathbf{u}_{real} \times \mathbf{v}_{real} - \mathbf{u}_{imaginary} \times \mathbf{v}_{imaginary} \\ \mathbf{u}_{real} \times \mathbf{v}_{imaginary} + \mathbf{u}_{imaginary} \times \mathbf{v}_{real} \end{bmatrix}$$

The *circular convolution* operation can also be implemented in this framework, but with consideration that the pairs of numbers are permuted together. This can be implemented with a circulant matrix $\mathbf{W}$ with size $(N/2, N/2)$:

$$\mathbf{w} \circledast \mathbf{u} = \begin{bmatrix} \mathbf{W} & \mathbf{0} \\ \mathbf{0} & \mathbf{W} \end{bmatrix} \mathbf{u}$$

```python
from __future__ import division
import numpy as np
import scipy.special

def ncdf(z):
    return 0.5 * (1 + scipy.special.erf(z/2**0.5))

def p_correct_snr(M, N=10000, D=27, ares=2000):
    p = np.zeros((ares-1, len(M)))
    for iM,Mval in enumerate(M):
        s = (N / Mval)**0.5
        # span the Hit distribution up to 8 standard deviations
        av = np.linspace(s - 8, s + 8, ares)
        # the discretized gaussian of h_d'
        pdf_hdp = ncdf(av[1:])-ncdf(av[:-1])
        # the discretized cumulative gaussian of h_d
        cdf_hd = ncdf(np.mean(np.vstack((av[1:]+s, av[:-1]+s)),
            axis=0))
        p[:, iM] = pdf_hdp * cdf_h ** (D-1)
    return np.sum(p, axis=0) # integrate over av
```

Figure 14: **Numeric algorithm for accuracy integral.**

The *superposition* ($+$) is the same, and *similarity* ($\cdot$) functions is defined for complex numbers as simply:

$$\mathbf{u} \cdot \mathbf{v} = \mathbf{u}_{real} \cdot \mathbf{v}_{real} + \mathbf{u}_{imaginary} \cdot \mathbf{v}_{imaginary}$$

which is the real part of the conjugate dot product, $Re(\mathbf{u}^\top \mathbf{v}^*)$.

Either circular convolution or element-wise multiplication can be used to implement binding in FHRR, and trajectory association can be performed to encode the letter sequence with either operation:

$$\mathbf{x}(M) = \sum \mathbf{w}^{M-m} \times \mathbf{\Phi a}(m) \text{ or}$$
$$\mathbf{x}(M) = \sum \mathbf{w}^{\circledast(M-m)} \circledast \mathbf{\Phi a}(m)$$

where $\mathbf{w}^{\circledast m}$ means circular convolutional exponentiation (e.g. $\mathbf{w}^{\circledast 2} = \mathbf{w} \circledast \mathbf{w}$).

## 4.2 Accuracy of retrieval from superpositions

### 4.2.1 Comparsion of approximations for the high-fidelity regime

We compared each step of the high-fidelity approximation (Results 2.2.2) to the true numerically evaluated integral, to understand which regimes the approximations were valid (Fig. 15B).

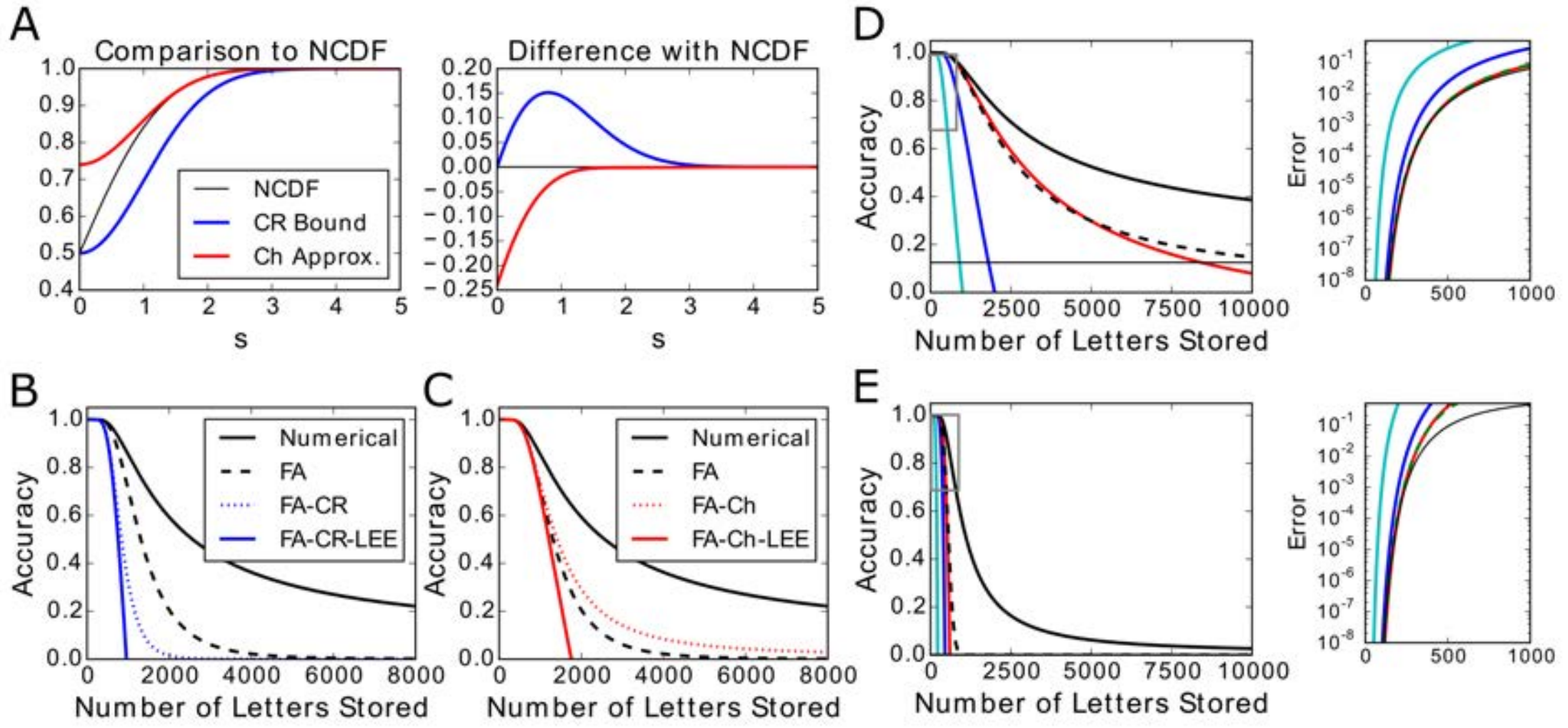

Figure 15: **Comparison of different methods to approximate the retrieval accuracy.** A. The Chernoff-Rubin (CR) (Chernoff, 1952) lower bound (blue) and the Chang et al. (2011) approximation (red) to compute the normalized cumulative density function (NCDF; black) analytically. The Chang et al. (2011) approximation becomes tight faster in the high-fidelity regime, but is not a lower bound. B. Differences between the three methods of approximations and the numerically evaluated $p_{corr}$ integral (black line). The factorial approximation (dashed black line) still requires numerical evaluation of the NCDF. Adding the CR lower-bound (dashed blue) and further the local-error expansion the high-fidelity regime can still be described well but the low-fidelity regime cannot be captured. C. Same as B, but using the Chang et al. (2011) approximation to the NCDF. D. Accuracy curve and approximations for $D = 8$. E. $D = 1024$. Right panels in D and E are zoom in's into the high-fideltiy regime (marked by gray box insets in the left panels).

We compare the CR bound and the Chang et al. (2011) approximation to the numerically evaluated Normal cdf $\Phi$ and see that the CR lower bound does not get tight until multiple standard deviations into the very high-fidelity regime (Fig. 15A).

In Fig. 15D, E, we see that while the approximations given are not strictly lower bounds, they are typically below the numerically evaluated accuracy. The Chang approximation can over-estimate the performance, however, in the high-fidelity regime when $D$ is large.

### 4.2.2 Previous theories of the high-fidelity regime

The capacity theory derived here is similar to, but slightly different from the analysis of Plate (2003), which builds from work done in Plate (1994). Plate (2003) frames the question: “What is the probability that I can correctly decode all $M$ tokens stored, each of which are taken from the full set of $D$ possibilities without replacement?” This is a slightly different problem, because this particular version of Plate (2003)’s anlaysis does not use trajectory association to store copies of the same token in different addresses. Thus $M$ is always less than $D$, the $M$ tokens are all unique, and there is a difference in the sampling of the tokens between our analysis frameworks.

Nonetheless, these can be translated to a roughly equivalent framework given that $D$ is relatively large compared to $M$. Plate (2003) derives the hit $p(h_{d'})$ and reject $p(h_d)$ distributions in the same manner as presented in our analysis, as well as uses a threshold to pose the probability problem:

$$p_{all-corr} = p(h_{d'} > \theta)^M p(h_d < \theta)^{D-M} \tag{69}$$

This can be interpreted as: the probability of reading all $M$ tokens correctly ($p_{corr-all}$) is the probability that the dot product of the true token is larger than threshold for all $M$ stored tokens ($p(h_{d'} > \theta)^M$) *and* that the dot product is below threshold for all $D - M$ remaining distractor tokens ($p(h_d < \theta)^{D-M}$).

In our framework, the probability of correctly reading out an individual symbol from the $M$ items stored in memory is independent for all $M$ items. This is (12), and to alter the equation to output the probability of reading all $M$ tokens correctly, then simply raise $p_{corr}$ to te $M$th power:

$$p_{all-corr} = [p_{corr}]^M = \left[\int_{\theta}^{\infty} \frac{dh}{\sqrt{2\pi}} e^{\frac{-h^2}{2}} \left[\Phi\left(h+s\right)\right]^{D-1}\right]^M \tag{70}$$

In Figure 16, we compare our theory to Plate’s by computing $p_{all-corr}$ given various different parameters of $N$, $M$, and $D$. We show that Plate (2003)’s framework comparatively underestimates the capacity of hypervectors. There is slight discrepancy in our analysis frameworks, because of how the tokens are decoded from memory. In our analysis framework, we take the maximum dot product as the decoded symbol, and there are instances that can be correctly classified that Plate (2003)’s probability statement (69) would consider as incorrect. For instance, the true symbol and a distractor symbol can both have dot products above threshold and the correct symbol can still be decoded as long as the true symbol’s dot product is larger than the distractor. However, this scenario would be classified as incorrect by (69).

Plate next uses an approximation to derive a linear relationship describing the accuracy. Citing Abramowitz et al. (1965), he writes:

$$erfc(x) < \frac{1}{x\sqrt{\pi}} e^{-x^2}$$

This approximation allows Plate to estimate the linear relationship between $N$, $M$,

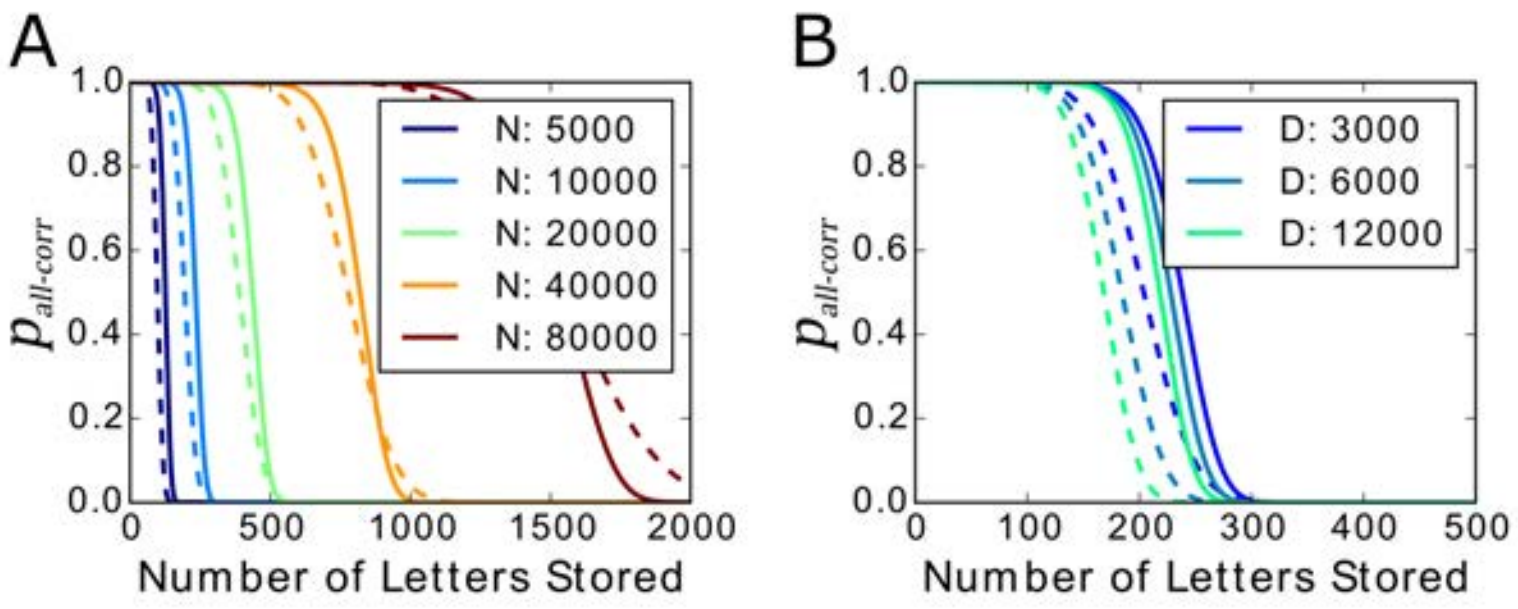

Figure 16: **Comparison with the theory in Plate (2003).** A. Plate (2003) derived $p_{all-corr} = p_{corr}^M$, plotted in dashed lines for different values of $N$ with $D$ fixed at 4096. The new theory in solid lines. B. Plate (2003)'s theory in dashed lines with different values of $D$ and fixed $N$. The new theory in solid lines.

$\log D$, and $\epsilon$. Arriving at:

$$N < 8M \log\left(\frac{D}{\epsilon}\right)$$

The FA-CR-LEE approximation only differs by a factor of 2, because of the slightly different choice we made to approximate the cumulative Gaussian as well as the different set-up for the problem.

Subsequent work by Gallant and Okaywe (2013) proposed an alternative VSA framework, which used a matrix as a binding mechanism. Based on their framework, they too in analogous fashion to Plate (2003) derived an approximation to the capacity of vector-symbols in superposition. Their derivation takes the high-fidelity factorial approximation as the starting point, and utilizes $e^{-x}$ as the bound on the tail of the normal distribution. This work is very similar to the derivation presented in this paper, but we add more rigor and derive a tighter high-fidelity approximation utilizing the Chernoff-Rubin bound and the updated approximation by Chang et al. (2011).

## 4.3 Derivations for analog Gaussian inputs

### 4.3.1 Analytic capacity bounds for Gaussian analog input

In Results 2.5.1, we derived the memory capacity for Gaussian inputs. Numerically, we showed that the equations suggest the memory capacity saturates to $1/(2\log 2)$ bits per neuron. It is possible to derive these capacity bounds analytically.

The total information is determined by $r(K)$ (57):

$$r(K) = \lambda^{2K} \frac{N(1-\lambda^2)}{D(1-\lambda^{2M})(1+\sigma_\eta^2/(DV_\Phi))} \tag{57}$$

and for Gaussian variables it can be computed as (58):

$$I_{total} = \frac{D}{2} \sum_{K}^{M} \log_2 \left( r(K) + 1 \right) \tag{58}$$

Inserting (57) into (58) we obtain:

$$I_{total} = \frac{D}{2} \sum_{K}^{M} \log_2 \left( r(K) + 1 \right) = \frac{D}{2} \log_2 \left( \prod_{K=1}^{M} \left( r(K) + 1 \right) \right) \tag{71}$$

The definition of the *$q$-Pochhammer symbol* or *shifted factorial*:

$$(a; q)_M := \prod_{K=0}^{M-1} (1 - aq^K) \tag{72}$$

yields a more compact expression (59):

$$I_{total} = \frac{D}{2} \log_2 \left( (-b_M q; q)_M \right) \tag{59}$$

with $q := \lambda^2$ and $b_M := \frac{N(1-q)}{D(1-q^M)(1+\sigma_\eta^2/(DV_{\mathbf{\Phi}}))}$.

The approximation of the logarithm of the $q$-Pochhammer symbol for $|b_M q| < 1$ will now be useful (Zhang, 2013):

$$\begin{aligned} \log\left( (b_M q; q)_\infty \right) &= \frac{1}{2} \log(1 - b_M q) - \frac{\tau}{2} Li_2(b_M q) - \frac{1}{6\tau} \frac{b_M q}{1 - b_M q} + O(1/\tau^3) \\ &= -\frac{\tau}{2} Li_2(b_M q) + \frac{1}{2} \log(1 - b_M q) + O(1/\tau) \end{aligned} \tag{73}$$

where $\tau := -2/\log(q)$ is the signal decay time constant for a given $q$ and $Li_2(x)$ is the dilogarithm function. Note that for $q$ close to one and large decay time constant the first term in (73) becomes the leading term.

For the case $M \to \infty$ and any $N$ we can lead $q$ so close to one that $b_\infty q = \epsilon$ becomes very small. This is accomplished by $q = 1 - \frac{\epsilon D(1+\sigma_\eta^2/(DV_{\mathbf{\Phi}}))}{qN}$ and, equivalently, $\tau = \frac{2Nq}{\epsilon D(1+\sigma_\eta^2/(DV_{\mathbf{\Phi}}))}$. With this setting we can apply approximation (73) to compute the asymptotic information capacity. Neglecting non-leading terms:

$$\begin{aligned} \frac{I_{total}}{N} &= \frac{D}{2N} \log_2 \left( (-b_\infty q; q)_\infty \right) \\ &= -\frac{D\tau}{4N \log(2)} Li_2(\epsilon) = \frac{q}{2\log(2)(1 + \sigma_\eta^2/(DV_{\mathbf{\Phi}}))} \\ &\approx_{q \to 1} \frac{0.72}{1 + \sigma_\eta^2/(DV_{\mathbf{\Phi}})} \end{aligned} \tag{74}$$

Equation (74) uses the result for the polylogarithm: $\lim_{|z| \to 0} Li_s(z) = z$.

For the case of finite $M$ and $D \propto N$, the identity $(a;q)_n = \frac{(a;q)_\infty}{(aq^n;q)_\infty}$ is useful, in combination with approximation (73). We consider $D = N/\alpha$ with $\alpha$, so that $b_M q = \frac{\alpha(1-q)q}{(1-q^M)(1+\sigma_\eta^2/(DV_{\mathbf{\Phi}}))} = \epsilon$ becomes very small. Further we set $\tau = -2/\log(q)$:

$$
\begin{aligned}
\frac{I_{total}}{N} &= \frac{D}{2N\log(2)}\left[\log\left((-b_M q;q)_\infty\right) - \log\left((-b_M q^{M+1};q)_\infty\right)\right] \\
&= \frac{\tau}{4\log(2)\alpha}\left[-Li_2\left(\epsilon\right) + Li_2\left(\epsilon q^M\right)\right] + O(\epsilon) + O(1/\tau) \\
&= \frac{(1-q)q(1-q^M)\tau}{4\log(2)(1-q^M)(1+\sigma_\eta^2/(DV_{\mathbf{\Phi}}))} + O(\epsilon) + O(1/\tau) \\
&= \frac{1}{2\log(2)(1+\sigma_\eta^2/(DV_{\mathbf{\Phi}}))} \times \frac{(1-q)q}{-\log(q))} + O(\epsilon) + O(1/\tau) \\
&\approx_{q\to 1} \frac{0.72}{1+\sigma_\eta^2/(DV_{\mathbf{\Phi}})}
\end{aligned} \tag{75}
$$

Thus, in both cases we find the same asymptotic value for the information capacity.

#### 4.3.2 Fixed fidelity retrieval optimization for memory buffer

With the relation between $r$, $K$, and $\tau$, we can find the $\tau_{opt}$ that maximizes $K^*$ such that $r(K) \geq r^* \; \forall K \leq K^*$. Beginning with the SNR (57):

$$
r(K) = \lambda^{2K}\frac{N(1-\lambda^2)}{D(1+\sigma_\eta^2/(DV_{\mathbf{\Phi}}))} = e^{-2K/\tau}\frac{N(1-e^{-2/\tau})}{D(1+\sigma_\eta^2/(DV_{\mathbf{\Phi}}))} \tag{57}
$$

Setting $r(K^*) = r^*$, and solving for $K^*$ gives:

$$
K^* = \frac{-\tau_{opt}}{2}\log\left(\frac{Dr^*(1+\sigma_\eta^2/(DV_{\mathbf{\Phi}}))}{N(1-e^{-2/\tau_{opt}})}\right) \tag{76}
$$

Taking the derivative $dK^*/d\tau_{opt}$ and setting to 0:

$$
-\frac{1}{2}\log\left(\frac{Dr^*(1+\sigma_\eta^2/(DV_{\mathbf{\Phi}}))}{N(1-e^{-2/\tau_{opt}})}\right) - \frac{e^{-2/\tau_{opt}}}{(1-e^{-2/\tau_{opt}})\tau_{opt}} = 0 \tag{77}
$$

For moderately large $\tau_{opt}$ the second term asymptotes to $\frac{1}{2}$, giving:

$$
\begin{aligned}
-1 &= \log\left(\frac{Dr^*(1+\sigma_\eta^2/(DV_{\mathbf{\Phi}}))}{N(1-e^{-2/\tau_{opt}})}\right) \\
e^{-1} &= \frac{Dr^*(1+\sigma_\eta^2/(DV_{\mathbf{\Phi}}))}{N(1-e^{-2/\tau_{opt}})} \\
\tau_{opt} &= \frac{-2}{\log\left(1-\frac{eDr^*(1+\sigma_\eta^2/(DV_{\mathbf{\Phi}}))}{N}\right)} \\
\frac{\tau_{opt}}{N} &= \frac{2}{eDr^*(1+\sigma_\eta^2/(DV_{\mathbf{\Phi}}))}
\end{aligned} \tag{78}
$$

From the first line of (78) and equation (76), it is easy to see that $K^* = \tau_{opt}/2$.

The information per neuron retrieved with a certain SNR criterion $r^*$ is then given by:

$$\frac{I^*(r^*)}{N} = \frac{D}{2N} \sum_{K=1}^{K^*} \log_2 \left(r(K) + 1\right) = \frac{D}{2N \log(2)} \log \left( (b_\infty q; q)_{\tau_{opt}/2} \right) \tag{79}$$

with (78), $q = e^{-2/\tau_{opt}}$ and $b_\infty = er^*$.

For small $r^*$ we can estimate:

$$\begin{aligned} \frac{I^*(r^* \to 0)}{N} &= \frac{D\tau_{opt}}{4N\log(2)} \left[ Li_2(b_\infty q) - Li_2(b_\infty q^{\tau_{opt}/2+1}) \right] \\ &= \frac{D\tau_{opt} b_\infty}{4N\log(2)} q(1 - q^{\tau_{opt}/2}) =_{r^* \to 0} \frac{1 - q^{\tau_{opt}/2}}{2\log(2)(1 + \sigma_\eta^2/(DV_{\mathbf{\Phi}}))} \\ &= \frac{1 - e^{-1}}{2\log(2)(1 + \sigma_\eta^2/(DV_{\mathbf{\Phi}}))} \approx \frac{0.46}{1 + \sigma_\eta^2/(DV_{\mathbf{\Phi}})} \end{aligned} \tag{80}$$

## 4.4 Capacity results with MMSE readout

### 4.4.1 Training procedure for the recurrent neural network

We used tensorflow to train a linear recurrent neural network at the sequence recall task. The parameter $K$ could be altered to train the network to output the symbol given to it in the sequence $K$ time steps in the history. However, larger $K$ requires deeper back propagation through time and becomes more expensive to compute and harder to learn. The training was based on optimizing the Energy function given by the cross-entropy between $\mathbf{a}(m - K)$ and $\hat{\mathbf{a}}(m - K)$. The accuracy was monitored by comparing the maximum value of the output histogram with the maximum of the input histogram.

We initialized the network to have a random Gaussian distributed encoding and decoding matrix ($\mathbf{\Phi}, \mathbf{V}(K)$) and a fixed random unitary recurrent weight matrix ($\mathbf{W}$). The random unitary matrix was formed by taking the unitary matrix from a QR decomposition of a random Gaussian matrix. Such a matrix maintains the energy of the network, and with a streaming input, the energy of the network grows over time. After a fixed number of steps ($M = 500$), the recurrent network was reset, where the activation of each neuron was set to 0. This erases the history of the input. Only outputs $K$ or more time steps after each reset were consider part of the energy function.

### 4.4.2 Comparison to DSR and normal networks in reservoir computing

White et al. (2004) describes the distributed shift register (DSR), a neural network which encodes an input sequence "spatially" by permuting the inputs down a chain of neurons. The final neuron in the chain, however, is not connected to any other post-synaptic neuron. This allows the network to store exactly $N$ numbers. Since the last neuron is not connected, the network remembers exactly the last $N$ most recent inputs, and the sequence length can be infinite. However, the network can be equated to our orthogonal networks with finite sequence length and a reset mechanism.

White et al. (2004) use a memory function that is the correlation between the input and decoded output to understand the information content of the DSR. This analysis extends to orthogonal networks with contracting weights, denoted *normal networks*. The correlation function for the DSR remains 1 until the lookback value $K$ exceeds $N$, when the correlation function drops to $0$. This has been taken as the upper limit on the information capacity for reservoir computing, and our results support this conclusion.

White et al. (2004) also derives an annealed approximation formula for the correlation function in normal networks. This approximation includes the MMSE readout, and corresponds to analog memory buffer networks considered in Results 2.6.2 (compare to Fig. 12). We compare their theory to the naive VSA readout performance and to the empirically measured performance of memory buffers with MMSE readout in Fig. 17. The curves in Fig. 17 matches those in White et al. (2004) Figure 2 (dot-dashed lines in Fig. 17). These curves are given by:

$$m(K) = \frac{\lambda^{2K} q}{1 + \lambda^{2K} q} \tag{81}$$

where $q$ satisfies:

$$1 = N^{-1} \sum_{K=0}^{\infty} \frac{\lambda^{2K} q}{1 + \lambda^{2K} q} + \frac{\sigma_\eta^2 q}{1 + \lambda^2} \tag{82}$$

These curves are generally more optimistic than the performance of naive VSA readout (Fig. 17, solid lines), except for larger time constants. However, the empirical performance of MMSE readout (Fig. 17, dashed lines) still highly outperforms both naive VSA readout and the White et al. (2004) theory. The MMSE readout performance, rather, approaches DSR-like performance, and nearly the full entropy of the neural space can be utilized if the time constant is appropriately optimized.

In our discrete framework, the discrete DSR would be considered a network with $D = 2$ (or with $D = 1$ and a detection threshold) and a sequence of binary inputs can be stored with $M$ increasing as high as $N$ but not higher. Because of the way this representation is constructed, there is no interference noise and the assumptions needed for our theory do not hold. However, the DSR can be reconsidered by focusing on the end result, rather than the time-evolution of the network. Ultimately, the DSR is a construction process that builds a binary representation for each possible binary input sequence and is equivalent to storing a single binary representation in the network. This is as if $M = 1$ and $D = 2^N$, where each possible input sequence corresponds to one of the $D$ code vectors. Thus, we are able to apply our $M = 1$ analysis to understand the information capacity of DSRs, and see how they are able to achieve the full entropy available.

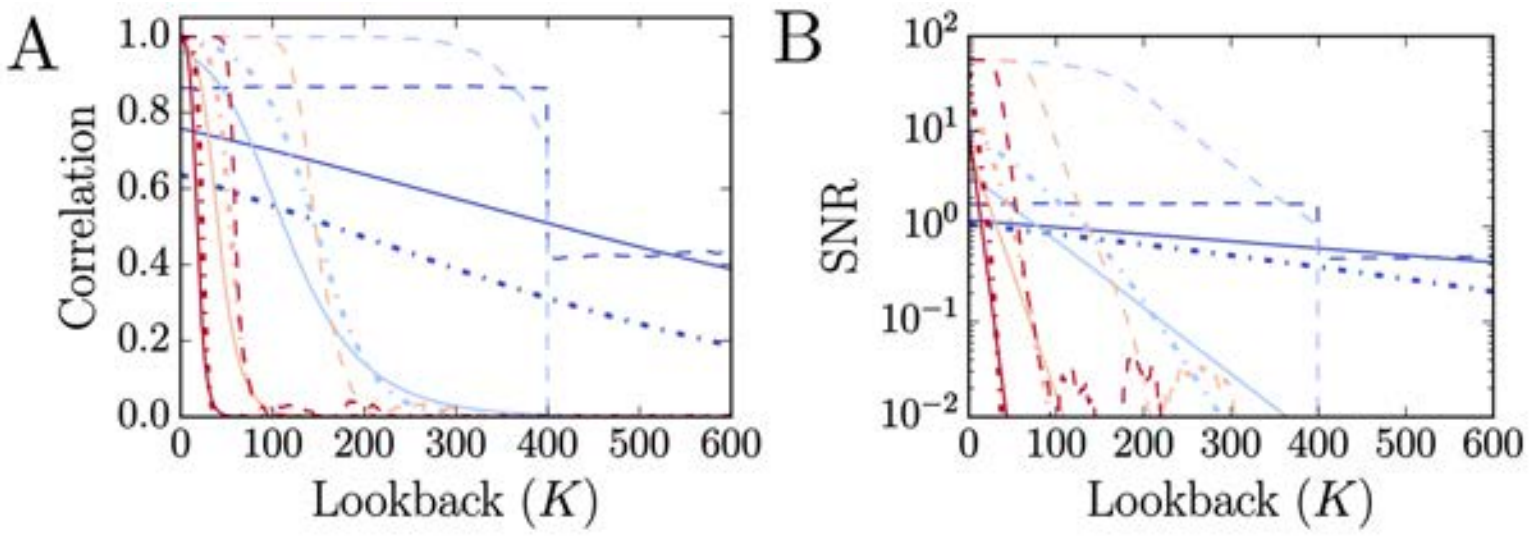

Figure 17: **Memory performance in normal networks** Comparison of naive VSA readout (solid lines), approximation from White et al. (2004) (dot-dashed lines), and empirical performance (dashed lines) for analog memory buffers with different time constants, $\lambda^2 = [0.7, 0.95, 0.99, 0.999]$ red to blue, $N = 400$, $D = 1$. These parameters taken from White et al. (2004).

#### 4.4.3 Randomized vector representations

In Results 2.3.5, we compared the memory capacity of superpositions to the memory capacity of the $M = 1$ case as $D \to 2^N$. As $D$ grows to a significant fraction of $2^N$, the crosstalk from superposition becomes overwhelming and the memory capacity is maximized for $M = 1$. The retrieval errors are then only due to collisions between the randomized codevectors and the accuracy $p_{corr}^{M=1}$ is given by (39). Fig. 18A, shows the accuracy for $M = 1$ as $D$ grows to $2^N$ with a randomly constructed codebook for different (smaller) values of $N$ – for large $N$ the numerical evaluation of (39) is difficult. As $N$ grows, the accuracy remains perfect for an increasingly large fraction of the $2^N$ possible code vectors. However, at the point $D = 2^N$ the accuracy falls off to $(1 - 1/e)$, but this fall-off is sharper as $N$ grows larger. The information retrieved from the network also grows closer and closer to 1 bit per neuron as $N$ grows larger with $M = 1$ (Fig. 18B).

In Fig. 18B the capacity $I_{total}/N$ of the randomly constructed codebook for $M = 1$ was computed with the equation we developed for superposed codes (27). However, the nature of the retrieval errors is different for $M = 1$, rather than crosstalk, collisions of code vectors is the error source. By performing an exhaustive analysis of the collision structure of a particular random codebook, the error correction can be limited to actual collisions and the capacity of such a retrieval procedure is higher. The information transmitted when using the full knowledge of the collision structure is:

$$I_{total} = \sum_C p_C \log_2 \left( \frac{p_C D}{C + 1} \right) \tag{83}$$

For $D = 2^N$ and $N \to \infty$, the total information of a random vector symbol approaches

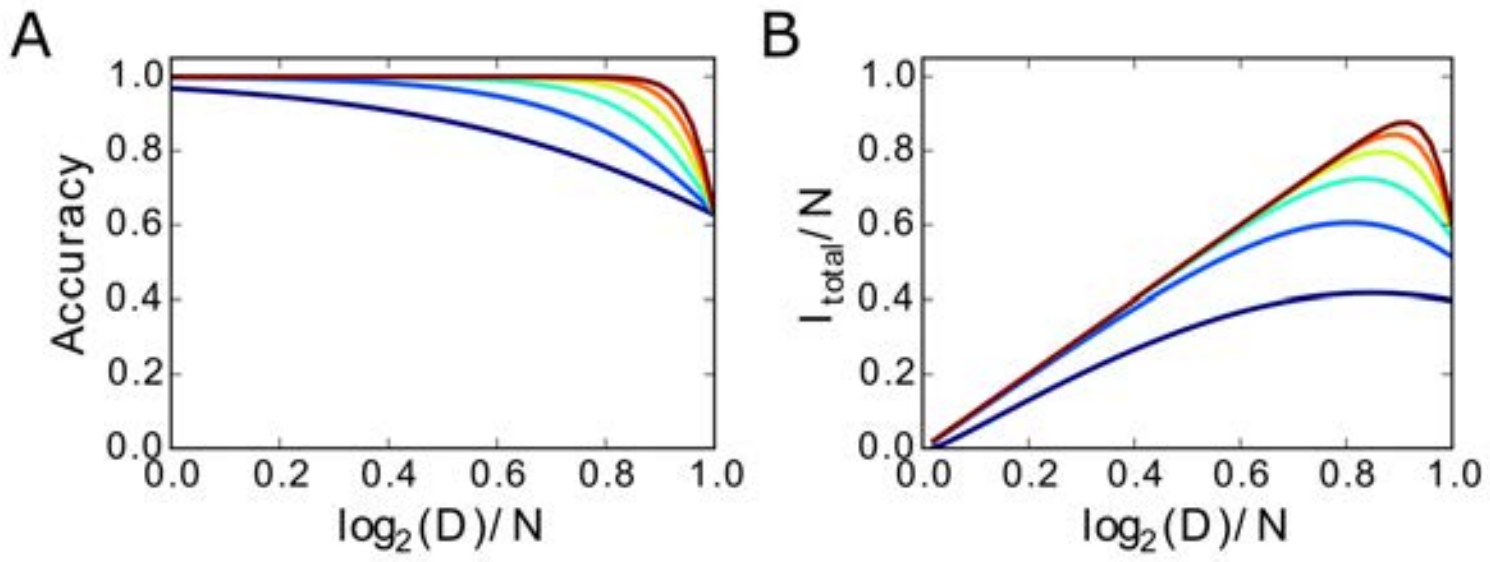

Figure 18: **Finite size effects on information capacity in discrete DSR's with randomized codes.** A. The accuracy $p_{corr}^{M=1}$ with increasing $N$. B. The retrieved information with increasing $N$.

1 bit per neuron:

$$\lim_{N\to\infty} \frac{1}{N} \sum_C p_C \left( N + \log_2 \left( \frac{p_C}{C+1} \right) \right) \to 1 \tag{84}$$

It is an interesting and somewhat surprising result in the context of DSRs that a random codebook yields asymptotically, for large $N$, the same capacity as a codebook in which collisions are eliminated by construction (White et al., 2004). But it has to be emphasized that a retrieval procedure, which uses the collision structure of the random codebook, is only necessary and advantageous for the $M = 1$ case. For superpositions, even with just two code vectors ($M = 2$), the alphabet size $D$ has to be drastically reduced to keep crosstalk under control and the probability of collisions between random code vectors becomes negligible.

### 4.4.4 Capacity with expected MMSE readout

Further following White et al. (2004), we were able to compute $\tilde{\mathbf{C}}$, the expected covariance matrix of MMSE readout, without any synthetic training data (Results 2.6). White et al. (2004) focuses on the memory buffer scenario where an infinite stream of inputs is given. They derive the result of MMSE readout in the buffer scenario, where they have an infinite input stream to act as training data. Their result is extended to higher $D$ by equation (67). This matches the performance of empirically trained memory buffers for both discrete and analog inputs.

We derive an analogous equation for the expected covariance matrix of linear reset memories (63). Computing $\mathbf{C}$ empirically requires training $R$ parallel neural networks with the same connectivity, but different input sequences. The covariance matrix $\mathbf{C}$ is

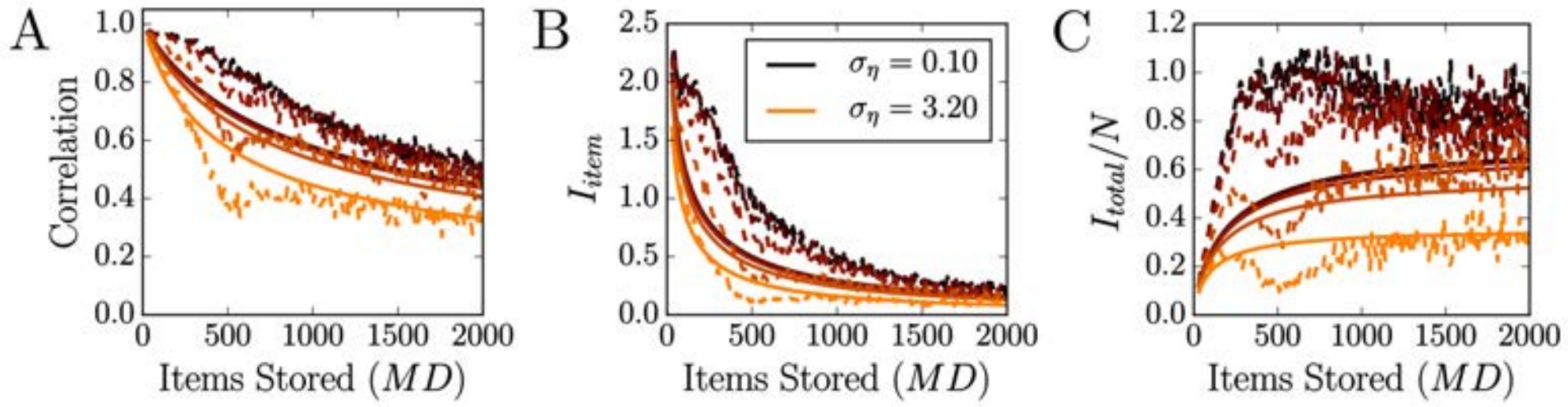

Figure 19: **Performance of analog reset memory with expected covariance matrix.** Compare to Fig. 11E-H.

given by:

$$\begin{aligned}\mathbf{C} =&\langle \mathbf{x}(M)\mathbf{x}(M)^\top \rangle_R \\ =&\frac{1}{R}\sum_r^R \left(\sum_{K_1}^M \mathbf{W}^{K_1}\mathbf{\Phi}\mathbf{a}(K_1;r) + \boldsymbol{\eta}(K_1;r)\right)\left(\sum_{K_2}^M \mathbf{W}^{K_2}\mathbf{\Phi}\mathbf{a}(K_2;r) + \boldsymbol{\eta}(K_2;r)\right)^\top \\ =&\sum_K^M \mathbf{W}^K\mathbf{\Phi}\left(\frac{1}{R}\sum_r^R \mathbf{a}(K;r)\mathbf{a}(K;r)^\top\right)\mathbf{\Phi}^\top\mathbf{W}^{-K} \\ &+\sum_{K_1}^M\sum_{K_2\neq K_1}^M \mathbf{W}^{K_1}\mathbf{\Phi}\left(\frac{1}{R}\sum_r^R \mathbf{a}(K_1;r)\mathbf{a}(K_2;r)\right)\mathbf{\Phi}^\top\mathbf{W}^{-K_2} \\ &+ M\sigma_\eta^2\mathbf{I}\end{aligned} \tag{85}$$

The covariance matrix is broken up into three parts, the diagonal term, the cross term, and the noise term. For $R \to \infty$, the diagonal term contains $\frac{1}{R}\sum_r^R \mathbf{a}(K;r)\mathbf{a}(K;r)^\top \to \mathbf{I}\sigma^2(\mathbf{a})$. The cross term converges to 0 if $\mu(\mathbf{a}) = 0$. This leaves (63):

$$\tilde{\mathbf{C}} = \langle \mathbf{x}(M)\mathbf{x}(M)^\top \rangle_\infty = M\sigma_\eta^2\mathbf{I} + \sum_{k=1}^M \mathbf{W}^k\mathbf{\Phi}\mathbf{\Phi}^\top\mathbf{W}^{-k} \tag{63}$$

This result does exceed the naive readout performance, however, it does not perform as well as an empirically trained MMSE network with finite $R$. Our simulations appear to be converging to the theory of equation (63), but second-order terms seem to play an important role in eliminating the crosstalk noise. These terms are small, but still significant even for very large $R$. The performance of readout with the expected covariance matrix is shown in Fig. 19 and can be compared to Fig. 11E-H.

# Acknowledgements

The authors would like to thank Pentti Kanerva, Bruno Olshausen, Guy Isely, Yubei Chen, Alex Anderson, Eric Weiss and the Redwood Center for Thoeretical Neuro-science for helpful discussions and contributions to the development of this work. This work was supported by the Intel Corporation (ISRA on Neuromorphic architectures for Mainstream Computing), and in part by the Swedish Research Council (grant no. 2015-04677)